\documentclass[11pt]{article}
\usepackage[preprint]{acl}
\usepackage{times}
\usepackage{latexsym}
\usepackage{iftex}
\ifPDFTeX
  \usepackage[T1]{fontenc}
  \usepackage[utf8]{inputenc}
\else
  \usepackage{fontspec}
\fi
\usepackage{microtype}
\usepackage{amsmath,amssymb}
\usepackage{graphicx}
\usepackage{booktabs,array,tabularx,colortbl,multirow}
\usepackage{placeins}
\usepackage{xurl}
\usepackage{pgfplots}
\pgfplotsset{compat=1.18}
\usepgfplotslibrary{groupplots}
\definecolor{oursrow}{rgb}{0.898,0.953,0.910}
\definecolor{acqcol}{HTML}{E07A2E}
\definecolor{forcol}{HTML}{3B5B7A}
\definecolor{retcol}{HTML}{1F8A9A}
\definecolor{ircol}{HTML}{1B8A4A}
\definecolor{hardcol}{HTML}{6DBE82}
\definecolor{boscol}{HTML}{8E6BB3}
\definecolor{dklcol}{HTML}{E377A0}
\definecolor{naivecol}{HTML}{D1453B}
\definecolor{loracol}{HTML}{9C755F}
\definecolor{srcol}{HTML}{4E79A7}
\definecolor{sdftcol}{HTML}{F28E2B}
\definecolor{dftcol}{HTML}{C9A227}
\definecolor{raftcol}{HTML}{3FA7A0}
\definecolor{othcol}{HTML}{B5B5B5}
\pgfplotsset{
  method classes/.style={scatter/classes={
    0={mark=x, black, mark size=2.6pt, thick},
    1={mark=*, naivecol, mark size=2.1pt},
    2={mark=square*, loracol, mark size=1.9pt},
    3={mark=triangle*, srcol, mark size=2.6pt},
    4={mark=triangle*, sdftcol, mark size=2.6pt, mark options={rotate=180}},
    5={mark=diamond*, dftcol, mark size=2.6pt},
    6={mark=pentagon*, raftcol, mark size=2.5pt},
    7={mark=*, dklcol, mark size=2.1pt},
    8={mark=square, boscol, mark size=2.1pt, thick},
    9={mark=diamond, hardcol, mark size=2.7pt, thick},
    10={mark=star, ircol, mark size=4pt, very thick}}},
  length classes/.style={scatter/classes={
    0={mark=*, othcol, mark size=1.5pt},
    1={mark=*, dklcol, mark size=1.9pt},
    2={mark=square, boscol, mark size=1.9pt, thick},
    3={mark=diamond, hardcol, mark size=2.4pt, thick},
    4={mark=star, ircol, mark size=3.2pt, thick},
    5={mark=*, naivecol, mark size=1.9pt}}},
}

\providecommand{\tightlist}{\setlength{\itemsep}{0pt}\setlength{\parskip}{0pt}}
\makeatletter
\newcommand{\fullrowcolor}[2]{\rlap{\color{#1}\vrule width #2 height \ht\@arstrutbox depth \dp\@arstrutbox}}
\makeatother

\title{InstructReplay: Replaying Assistant Turns to Preserve General Behavior During Factual Fine-Tuning}
\author{Xiaofeng Shi\thanks{Corresponding author: \texttt{xfshi@baai.ac.cn}.} \quad Qian Kou \quad Hua Zhou \\
Beijing Academy of Artificial Intelligence (BAAI)}
\hypersetup{pdftitle={InstructReplay: Replaying Assistant Turns to Preserve General Behavior During Factual Fine-Tuning},
pdfauthor={Xiaofeng Shi, Qian Kou, Hua Zhou},pdfsubject={arXiv preprint}}

\begin{document}
\maketitle

\begin{abstract}
Fine-tuning an instruction-tuned language model on new facts often erodes its general abilities. We find that standard fine-tuning lowers instruction-following, reasoning and coding benchmarks by 35 points on average, and that these losses track a collapse in answer length: fine-tuned models at least halve their responses, and the further a response drifts from the initial model's length, the more often it fails. We propose InstructReplay: the initial model answers a set of general instructions once, and during fine-tuning the model is kept close to the initial model on these answers. Across three model families and two domains, InstructReplay reduces the average decline on five general benchmarks from 28.1 to 2.4 points and keeps answer length close to the initial model's, while learning the new facts as well as standard fine-tuning. It balances learning and retention better than seven adaptation methods in all six settings. Controlled comparisons attribute the gain to replaying the model's own answers to general instructions rather than to the form of the loss.
\end{abstract}

\section{Introduction}\label{sec:intro}

\begin{figure*}[t]
\centering
\includegraphics[width=\textwidth]{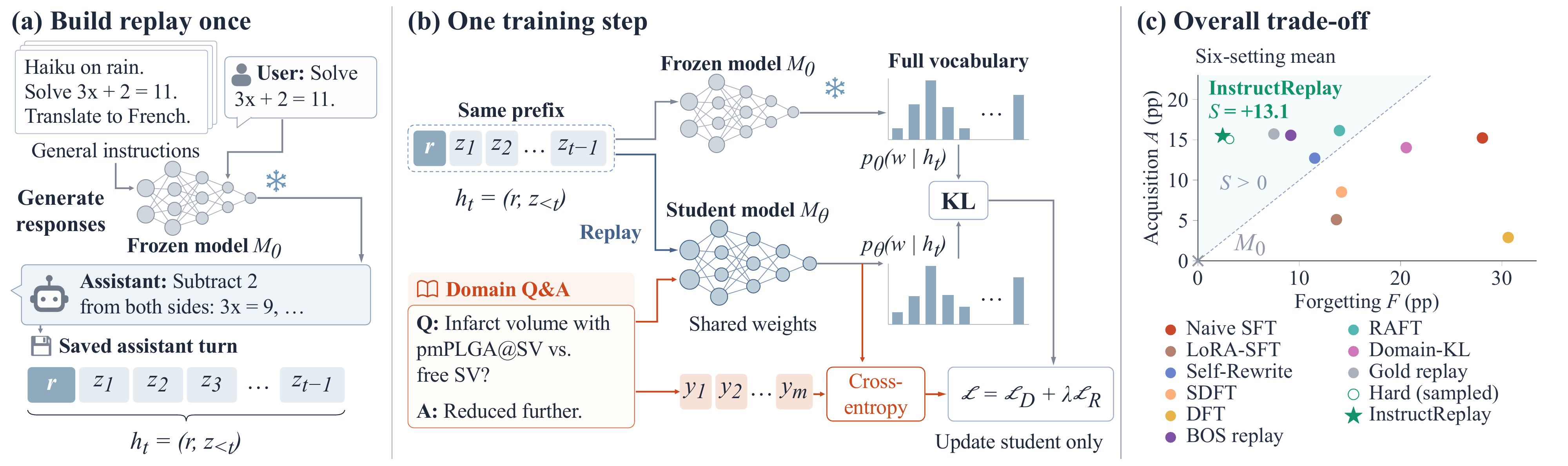}
\caption{Overview of InstructReplay. (a) Before training, the frozen initial model $M_0$ answers each general instruction once under its chat template, and these assistant turns form the replay buffer; the dataset's own answers are not used. (b) In each step the student matches $M_0$'s full next-token distribution at every saved prefix of the replayed turns while it learns domain answers with cross-entropy; the two losses share the student's weights. (c) Equal-weight mean acquisition $A$ and forgetting $F$ over six model--domain settings, combining the adaptation methods in Table~\ref{tab:results-1} and replay controls in Table~\ref{tab:where}. The green star marks InstructReplay; shading indicates $S=A-F>0$.}
\label{fig:overview}
\end{figure*}

Practitioners often adapt an instruction-tuned language model to a domain by fine-tuning it on question--answer pairs that state new facts. The facts are learned, but at a cost elsewhere. Fine-tuned on recent biomedical articles, SmolLM3-3B answers 21 points more of the new questions correctly, yet its instruction-following accuracy on IFEval falls from 71\% to 9\%. Fine-tuning is known to be an inefficient way of adding knowledge \citep{gekhman2024newknowledge,ovadia2024finetuning}. We study how to preserve general assistant behavior during closed-book factual fine-tuning.

Where does the damage occur? Standard fine-tuning sharply lowers accuracy on every benchmark that requires the model to generate an answer, from instruction following to mathematics, reasoning and code, yet barely changes MMLU, which we score by option likelihood. The adapted models also stop answering at length. Their responses shrink toward the short domain answers they were trained on, and the more a method shortens responses, the more accuracy it loses (\S\ref{sec:diagnosis}). The adapted models thus stop behaving as the assistants they were, even though likelihood-based probes change little.

Replaying an earlier model's predictions is a standard remedy for such losses \citep{li2016lwf,buzzega2020der}. A distillation penalty, however, constrains the model only at the prefixes where it is applied, so the choice of replay inputs decides which behavior is protected. Recent self-generated replay for instruction-tuned models samples from a beginning-of-sequence (BOS) token, outside the chat template \citep{marek2026forgetting}, and context-free synthetic data likewise avoids external prompts \citep{bansal2025contextfree}. Self-synthesized rehearsal does use instruction contexts, but it synthesizes and refines them and trains on the refined outputs \citep{huang2024ssr}. What has not been isolated is which ingredients of replay matter: \emph{where} the reference model is matched, \emph{whose} responses are replayed, and in what form.

To study this question, we introduce InstructReplay, a minimal form of instruction-context replay (Figure~\ref{fig:overview}). The initial model answers 9,600 general instructions once, and during domain training the model matches the frozen initial model's full next-token distributions on these saved responses. We compare it with seven domain-adaptation methods and with three controlled variants. The first keeps the objective but applies it elsewhere, to samples generated from a beginning-of-sequence token or to the domain answers themselves. The second matches sampled tokens instead of full distributions, and the third replays the instruction dataset's own answers instead of the model's. Our findings are:

\begin{itemize}\tightlist
\item \textbf{Factual fine-tuning mainly damages generative behavior.} Standard fine-tuning lowers the four generative benchmarks by 35 points on average and at least halves response length, while likelihood-scored MMLU barely moves (\S\ref{sec:diagnosis}).
\item \textbf{Instruction-context replay removes most forgetting while keeping acquisition on par with standard fine-tuning.} Replaying the model's own answers to general instructions reduces the average general decline from 28.1 to 2.4 points, and its acquisition stays within two points of standard fine-tuning in every setting (\S\ref{sec:main}).
\item \textbf{What is replayed matters more than how it is matched.} Moving the same constraint from the domain answers to unprompted samples and then to the model's own instruction responses reduces the average IFEval decline from 43 to 14 and then to about 3 points. Under sampled targets, replaying the dataset's answers to the same instructions instead of the model's own more than doubles forgetting, from 3.1 to 7.5 points, whereas soft and sampled targets perform similarly (\S\ref{sec:where}).
\end{itemize}

\section{Related Work}\label{sec:related}

\textbf{Learning new facts by fine-tuning.} Facts that a model does not already know are learned more slowly than known ones and, once learned, increase hallucination \citep{gekhman2024newknowledge}, and retrieval augmentation outperforms unsupervised fine-tuning for knowledge injection \citep{ovadia2024finetuning}. We take closed-book fine-tuning as given and study its cost to general behavior.

\textbf{Preserving earlier behavior.} Parameter-space penalties such as EWC protect weights deemed important for earlier tasks \citep{kirkpatrick2017ewc}; low-rank adaptation \citep{hu2021lora} learns less and forgets less than full fine-tuning \citep{biderman2024lora}. Function-space methods instead match an earlier model's outputs: Learning without Forgetting distills the old model on new-task inputs \citep{li2016lwf}, Dark Experience Replay stores logits alongside rehearsed examples \citep{buzzega2020der}, and label annealing distills the initial model on the fine-tuning data itself \citep{yang2024annealing}. Rehearsing a small share of earlier tasks preserves abilities in sequentially fine-tuned language models \citep{scialom2022continual}. Forgetting after fine-tuning often reflects a shift in how the model responds rather than a loss of capability \citep{kotha2024implicit,zheng2025spurious}, and updates driven by a model's own on-policy outputs forget less than supervised fine-tuning on external targets \citep{shenfeld2025razor,chen2026retaining}. Our Domain-KL baseline places the distillation term on domain answers, as label annealing does, and therefore isolates the choice of context.

\textbf{Choosing what to replay.} When the original training data are unavailable, replay must be constructed. Self-Synthesized Rehearsal generates and refines synthetic instructions \citep{huang2024ssr}; GeRe replays general text and compares label, logit and feature targets \citep{zhang2025gere}; \citet{bansal2025contextfree} train on context-free synthetic responses; and \citet{marek2026forgetting} apply frozen-reference forward KL to self-generated replay, sampled from BOS for instruction-tuned models. Section~\ref{sec:setup-baselines} defines our BOS replay control.

\textbf{Changing the domain supervision.} Another line keeps the data but changes the targets: Self-Distillation Fine-Tuning rewrites gold answers with the initial model \citep{yang2024sdft}, on-policy self-distillation trains on the model's own responses \citep{shenfeld2026sdft,agarwal2023gkd}, RAFT refines answers and adds adaptive distillation \citep{li2026raft}, and DFT reweights target tokens \citep{wu2025dft}. Both methods are called SDFT; we call the offline one \emph{Self-Rewrite}.

\section{Where Does Factual Fine-Tuning Hurt?}\label{sec:diagnosis}

We first characterize the damage that motivates our method, using standard fine-tuning (Naive SFT) in the setup of \S\ref{sec:setup}. Three instruction-tuned models are each fine-tuned separately on 9,600 question--answer pairs from two domains and evaluated on five general benchmarks against the unchanged initial model $M_0$.

\begin{table}[t]
\centering\small
\setlength{\tabcolsep}{2.4pt}
\begin{tabular}{@{}lrrr@{}}
\toprule
 & \multicolumn{1}{c}{$\Delta$ accuracy (pp)} & \multicolumn{2}{c}{Median words} \\
\cmidrule(lr){2-2}\cmidrule(lr){3-4}
Benchmark & mean [min, max] & $M_0$ & SFT \\
\midrule
\multicolumn{4}{@{}l}{\textit{Generated answers}} \\
IFEval & $-44.9$ [$-61.7$, $-33.3$] & 190 & 20 \\
MATH-500 & $-28.0$ [$-45.0$, $-5.0$] & 242 & 42 \\
BBH & $-36.0$ [$-70.0$, $-15.6$] & 122 & 54 \\
HumanEval+ & $-30.5$ [$-51.6$, $-6.2$] & 46 & 23 \\
\addlinespace[2pt]
\multicolumn{4}{@{}l}{\textit{Answer-option likelihood}} \\
MMLU & $-0.4$ [$-2.1$, $+2.8$] & -- & -- \\
\bottomrule
\end{tabular}
\caption{Standard fine-tuning on domain facts versus $M_0$, over six model--domain settings. Benchmarks that require a generated answer collapse and responses shorten; MMLU, scored by the likelihood of its answer options, barely moves and produces no response to measure. Median words (whitespace-separated) average the per-setting medians.}
\label{tab:diagnosis}
\end{table}

\textbf{The damage is concentrated in generation.} Table~\ref{tab:diagnosis} contrasts the four benchmarks that require the model to generate an answer with MMLU, which we score by the likelihood of the answer options. The generative benchmarks fall by 35 points on average, whereas MMLU stays within three points in every setting. On BIG-Bench Hard, the share of answers that the task-aware parser cannot extract rises from 6\% to 45\% (Appendix Table~\ref{tab:results-10}), so much of the loss comes from answers that no longer follow the task's answer format.

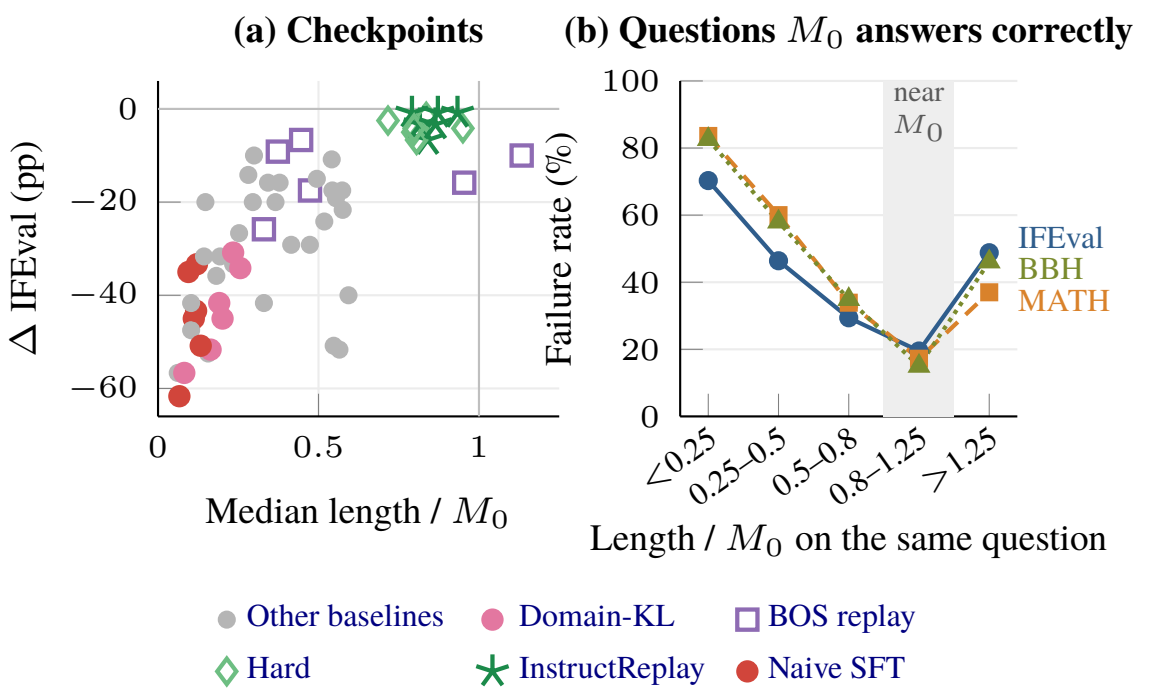
\begin{figure}[t]
\centering
\definecolor{benchif}{HTML}{2F5D8C}
\definecolor{benchmath}{HTML}{D9822B}
\definecolor{benchbbh}{HTML}{7A8B2E}
\begin{tikzpicture}
\begin{groupplot}[group style={group size=2 by 1, horizontal sep=0.8cm},
  width=0.55\columnwidth, height=3.8cm, label style={font=\scriptsize}, ticklabel style={font=\tiny},
  title style={font=\scriptsize\bfseries, yshift=-4pt}, axis lines*=left, ymajorgrids, grid style={gray!15}]
\nextgroupplot[title={(a) Checkpoints}, xmin=0, xmax=1.25, ymin=-66, ymax=6, xmajorgrids,
  xlabel={Median length / $M_0$}, ylabel={$\Delta$ IFEval (pp)}, ylabel style={yshift=-3pt}, length classes,
  legend to name=lenmainlegend, legend columns=3, legend cell align=left,
  legend style={font=\tiny, draw=none, /tikz/every even column/.append style={column sep=4pt}}]
\addplot[gray!50, thin, forget plot] coordinates {(1,-66) (1,6)};
\addplot[gray!50, thin, forget plot] coordinates {(0,0) (1.25,0)};
\addplot[scatter, only marks, scatter src=explicit symbolic] table[x=ratio, y=delta, meta=cls] {figures/data/length-IFEval.dat};
\legend{Other baselines, Domain-KL, BOS replay, Hard, InstructReplay, Naive SFT}
\nextgroupplot[title={(b) Questions $M_0$ answers correctly}, xmin=0.6, xmax=5.4, ymin=0, ymax=100, clip=false,
  width=0.495\columnwidth,
  xtick={1,2,3,4,5}, xticklabels={$<$0.25, 0.25--0.5, 0.5--0.8, 0.8--1.25, $>$1.25},
  xticklabel style={font=\tiny, rotate=35, anchor=north east, inner sep=1pt},
  xlabel={Length / $M_0$ on the same question}, xlabel style={yshift=2pt}, ylabel={Failure rate (\%)}, ylabel style={yshift=-3pt}]
\fill[gray!13] (axis cs:3.5,0) rectangle (axis cs:4.5,100);
\node[font=\fontsize{5.5}{6}\selectfont, text=gray!70!black, align=center] at (axis cs:4,90) {near\\ $M_0$};
\addplot[thick, color=benchif, mark=*, mark size=1.4pt] table[x=bin, y=IFEval] {figures/data/item-length.dat};
\addplot[thick, densely dashed, color=benchmath, mark=square*, mark size=1.3pt, mark options={solid}] table[x=bin, y=MATH] {figures/data/item-length.dat};
\addplot[thick, densely dotted, color=benchbbh, mark=triangle*, mark size=1.7pt, mark options={solid}] table[x=bin, y=BBH] {figures/data/item-length.dat};
\node[font=\tiny, text=benchbbh, anchor=west] at (axis cs:5.12,44.0) {BBH};
\node[font=\tiny, text=benchif, anchor=west] at (axis cs:5.12,53.5) {IFEval};
\node[font=\tiny, text=benchmath, anchor=west] at (axis cs:5.12,34.5) {MATH};
\end{groupplot}
\end{tikzpicture}

\ref{lenmainlegend}
\caption{(a) IFEval accuracy change versus median response length relative to $M_0$ for the 60 adapted checkpoints (Spearman $\rho=0.77$; Appendix Figure~\ref{fig:length} adds MATH-500 and BBH). (b) Failure rate on questions that $M_0$ answers correctly, by the length of each response relative to $M_0$'s on the same question, pooled over the same checkpoints (Appendix Table~\ref{tab:item-length}).}
\label{fig:length-main}
\end{figure}

\begin{table}[t]
\centering\small
\setlength{\tabcolsep}{4pt}
\begin{tabularx}{\columnwidth}{@{}>{\raggedright\arraybackslash}p{0.25\columnwidth}>{\raggedright\arraybackslash}X@{}}
\toprule
\multicolumn{2}{@{}>{\raggedright\arraybackslash}p{\columnwidth}@{}}{\textit{Prompt:} Write a letter to a friend in all lowercase letters ask them to go and vote.} \\
\midrule
$M_0$\newline 155 words, \checkmark & dear [friend's name], i hope this letter finds you well and in good spirits. i wanted to take a moment to reach out and ask you to consider casting your vote in the upcoming election. \ldots \\
\addlinespace[3pt]
Naive SFT\newline 8 words, $\times$ & \textcolor{red}{Please} go and vote in the upcoming election. \\
\addlinespace[3pt]
InstructReplay\newline 195 words, \checkmark & dear friend, i hope this letter finds you well. i wanted to take a moment to ask you to consider going and voting in the upcoming election. \ldots \\
\bottomrule
\end{tabularx}
\caption{An IFEval prompt and SmolLM3's responses on PMC (truncated; \checkmark/$\times$: constraint met or not). After standard fine-tuning the model answers in one sentence, like its short training answers, and breaks the lowercase constraint (red); InstructReplay keeps $M_0$'s letter.}
\label{tab:example}
\end{table}

\textbf{Responses become short, and shortening predicts the loss.} Median responses to IFEval prompts shrink from 190 to 20 words, and MATH-500 solutions from 242 to 42; Table~\ref{tab:example} shows an example in which a request for a letter receives a single sentence. This matches the training data: domain answers average about 30 tokens, while the initial models' answers to general instructions average about 340. Across the 60 adapted models, a model's median response length relative to its $M_0$ is strongly rank-correlated with its accuracy change (Figure~\ref{fig:length-main}a). Spearman's $\rho$ is 0.77 on IFEval, 0.81 on MATH-500 and 0.84 on BIG-Bench Hard (Appendix~\ref{app:length}).

\textbf{The losses are not only word-count failures.} Naive SFT's IFEval pass rate falls by 11 points on length constraints but by 55 points or more on constraints that require structure or specific content, such as formats, required openings and endings, and combined constraints (Appendix Table~\ref{tab:ifeval-categories}). Among questions that $M_0$ answers correctly, the failure rate of adapted models is lowest, below 20\%, when a response stays within 25\% of $M_0$'s length, and far higher when it is much shorter or much longer (Figure~\ref{fig:length-main}b). What predicts failure is thus departure from $M_0$'s response, of which shortening is the most common form.

\textbf{Hypothesis.} Fitting short domain answers shifts the model's behavior in assistant turns toward terse, direct answers, which breaks instruction following and chain-of-thought formatting; knowledge probed by likelihood is affected far less. If the hypothesis holds, a retention penalty should be most effective when it is applied in the contexts where assistant behavior is exercised.

\section{InstructReplay}\label{sec:method}

\subsection{Objective}\label{sec:objective}

Let $M_0$ be the initial instruction-tuned model with next-token distribution $p_0$, and $p_\theta$ the distribution of its trainable copy. The domain data $D=\{(x_i,y_i)\}$ contain questions and source-supported answers. Before training, $M_0$ samples one response $z_j$ to each prompt $r_j$ of a general instruction collection under its native chat template, giving a fixed replay buffer $\mathcal B_R=\{(r_j,z_j)\}_{j=1}^{N_R}$. Let $T(\cdot)$ denote the response-token positions of a sequence. The domain loss is the per-sequence mean cross-entropy
\[
\mathcal L_D=-\mathbb E_D\Bigl[\tfrac{1}{|T(y)|}\textstyle\sum_{t\in T(y)}\log p_\theta(y_t\mid x,y_{<t})\Bigr],
\]
and the replay loss matches $M_0$'s full-vocabulary distribution at every saved prefix $h_t=(r,z_{<t})$:
\begin{align*}
\mathcal L_R&=\mathbb E_{\mathcal B_R}\Bigl[\tfrac{1}{|T(z)|}\textstyle\sum_{t\in T(z)}\ell(h_t)\Bigr],\\
\ell(h)&=D_{\mathrm{KL}}\bigl(p_0(\cdot\mid h)\,\|\,p_\theta(\cdot\mid h)\bigr).
\end{align*}
We minimize $\mathcal L_D+\lambda\mathcal L_R$ with equal numbers of domain and replay sequences in every update. The reference is frozen, and inference uses only the adapted model. Appendix~\ref{app:loss} gives the local properties of $\mathcal L_R$.

\subsection{Why instruction contexts?}\label{sec:why}

Forward KL at a fixed prefix $h$ has logit gradient $p_\theta(\cdot\mid h)-p_0(\cdot\mid h)$. It pulls the student toward the reference only at the prefixes in the replay set and affects other contexts only through shared parameters, so the choice of prefixes determines what is preserved. Near $\theta_0$, the replay loss is a quadratic penalty whose curvature is built only from the logit Jacobians at the replayed prefixes (Appendix~\ref{app:loss}). It resists parameter changes that alter the model's outputs in those contexts and leaves other directions unconstrained. Domain-answer prefixes, as in Domain-KL, lie in exactly the contexts that $\mathcal L_D$ is changing. BOS-conditioned samples are free-form continuations generated without the chat template, which an instruction-tuned model rarely encounters at inference. Responses to general instructions under the native template place the constraint on assistant turns spanning instruction following, reasoning, mathematics, code and translation. These are the contexts in which general benchmarks and users query the model, and in which \S\ref{sec:diagnosis} located the damage. Section~\ref{sec:where} tests this ordering, comparing instruction and BOS contexts at a matched response-token budget.

We use soft targets because they exert no pull where the student already agrees with the reference, whereas a sampled token yields a nonzero gradient at every prefix (Appendix~\ref{app:loss}). A token sampled from $M_0$ still gives an unbiased estimate of the forward-KL gradient, so the two targets differ in variance rather than in what they preserve, and we expect the choice of prefixes, not the form of the target, to matter most. Section~\ref{sec:where} compares soft and sampled-token targets at the same replay coefficient.

\section{Experimental Setup}\label{sec:setup}

\subsection{Models, domains and replay data}

We adapt SmolLM3-3B \citep{bakouch2025smollm3}, Qwen3-4B \citep{qwen2025qwen3} and Phi-4-mini-instruct \citep{microsoft2025phi4mini}; each domain adaptation starts from the unchanged checkpoint $M_0$. Each domain provides 9,600 training question--answer pairs and 512 evaluation questions that ask about taught facts in new wording, without a source passage. CommonCorpus energy/engineering \citep{pleias2024commoncorpus} draws on technical documents; PMC biomedical \citep{ncbi2026pmc} uses CC BY 4.0 articles published online between May and August 2026, after all three models were released. Appendix~\ref{app:data} describes construction, examples and the extent to which some questions cue part of the answer.

The replay collection holds 9,600 prompts from UltraData-SFT \citep{openbmb2026ultradata}, split equally across instruction following, academic tasks, mathematics, code and translation and selected without reference to model scores. Each model samples its own responses (temperature 1, at most 512 tokens); the dataset's own assistant answers are used only by the Gold replay control (\S\ref{sec:setup-baselines}).

\subsection{Measuring acquisition and forgetting}\label{sec:metrics}

\textbf{Acquisition.} Three judges from different model families (Qwen3.8-27B, Gemma-4-31B-it, GPT-OSS-20B) compare each greedy answer with a source-supported fact card; an answer is confirmed when at least two judges validly find all core facts supported. With $C$ and $C^0$ confirmations for the adapted model and $M_0$ out of $N_D=512$ questions, acquisition is $A=100(C-C^0)/N_D$. Unresolved and ineligible questions stay in the denominator; on CommonCorpus, a fixed fact-card rule makes 374 of the 512 questions eligible (Appendix~\ref{app:scoring}).

\textbf{Forgetting.} We measure general ability on fixed subsets of IFEval \citep{zhou2023ifeval}, MMLU \citep{hendrycks2020mmlu}, MATH-500 \citep{hendrycks2021math,lightman2023lets}, BBH \citep{suzgun2022bbh} and HumanEval+ \citep{liu2023evalplus}, using the same questions across models and methods. Subset sizes and selection procedures are detailed in Appendix~\ref{app:general}. With $G_k$ and $G_k^0$ the adapted and initial scores on benchmark $k$, forgetting is the mean decline $F=\frac{1}{5}\sum_k\max(0,G_k^0-G_k)$, and $W$ is the largest single decline. We summarize the trade-off by $S=A-F$, but report $A$ and $F$ throughout and show in \S\ref{sec:robust} that the conclusions hold for any weighting that puts forgetting at more than 0.31 times acquisition.

\subsection{Baselines and controls}\label{sec:setup-baselines}

\textbf{Domain-adaptation baselines.} Naive SFT and LoRA-SFT train on the gold answers; Self-Rewrite, SDFT and RAFT change the domain supervision; DFT reweights target tokens; and Domain-KL adds our KL term ($\lambda=0.25$) on domain-answer prefixes. Appendix~\ref{app:training} and Table~\ref{tab:results-11} give their implementations.

\textbf{Replay controls.} Three controls keep InstructReplay's $\lambda=0.25$, 1:1 ratio and schedule and change one component at a time. \emph{BOS replay} changes the context: it applies the KL to samples generated from a beginning-of-sequence token without the chat template, the sampling used by \citet{marek2026forgetting}. This comparison evaluates BOS sampling under our training protocol, not their full method; Appendix~\ref{app:replay} lists the configuration differences. \emph{Hard} changes the target: it keeps InstructReplay's saved responses but replaces the KL with next-token cross-entropy. Both match InstructReplay's number of response tokens sequence by sequence. \emph{Gold replay} changes the author of the replayed answers: it keeps Hard's prompts and loss but trains on the collection's own assistant answers, truncated to 512 tokens.

\textbf{Training.} Every method runs 300 optimizer steps with 32 domain sequences per step, so all methods see the same domain examples and are evaluated at the same step; replay methods add 32 replay sequences per step. Full-parameter methods use a constant learning rate of $10^{-5}$ and LoRA uses $10^{-4}$ (Appendix~\ref{app:training}). The replay ratio, $\lambda=0.25$ and the schedule were tuned on CommonCorpus with SmolLM3 and Qwen3 and transferred unchanged to Phi-4-mini and to PMC.

\subsection{Study design and uncertainty}\label{sec:design}

PMC tests the same recipe, unchanged, on a second domain. For question-sampling uncertainty we use a paired cluster bootstrap that resamples domain questions by shared source and general-benchmark questions within benchmark strata, with Bonferroni correction over all 36 pairs among the eight methods and $M_0$ in each setting (Appendix~\ref{app:stats}).

\section{Results}\label{sec:results}

\subsection{Main comparison}\label{sec:main}

\begin{table*}[t]
\centering\small
\setlength{\tabcolsep}{4pt}
\begin{tabular*}{\textwidth}{@{\extracolsep{\fill}}lrrrrrrrr@{}}
\toprule
 & \multicolumn{3}{c}{CommonCorpus: $S$} & \multicolumn{3}{c}{PMC: $S$} & \multicolumn{2}{c}{Mean of six} \\
\cmidrule(lr){2-4}\cmidrule(lr){5-7}\cmidrule(lr){8-9}
Method & SmolLM3 & Qwen3 & Phi-4-mini & SmolLM3 & Qwen3 & Phi-4-mini & $A\uparrow$ & $F\downarrow$ \\
\midrule
Naive SFT & $-17.49$ & $-11.54$ & $-13.94$ & $-22.07$ & $-1.17$ & $-10.83$ & 15.23 & 28.07 \\
LoRA-SFT & $-16.02$ & $-8.63$ & $-7.88$ & $-9.90$ & $-0.87$ & $-8.04$ & 5.11 & 13.67 \\
Self-Rewrite & \underline{$-5.83$} & \underline{6.82} & \underline{2.06} & $-11.76$ & 8.44 & 7.50 & 12.73 & \underline{11.52} \\
SDFT & $-10.82$ & $-0.72$ & $-0.88$ & $-10.65$ & $-9.29$ & $-1.48$ & 8.53 & 14.17 \\
DFT & $-30.55$ & $-20.04$ & $-24.10$ & $-42.25$ & $-26.24$ & $-23.15$ & 2.90 & 30.62 \\
RAFT & $-10.23$ & 6.80 & 0.80 & \underline{$-5.91$} & \underline{9.79} & \underline{11.86} & \textbf{16.15} & 13.96 \\
Domain-KL & $-13.83$ & $-6.60$ & $-4.79$ & $-15.31$ & $-7.66$ & 9.04 & 14.03 & 20.55 \\
\fullrowcolor{oursrow}{\textwidth}InstructReplay & \textbf{1.01} & \textbf{8.56} & \textbf{7.22} & \textbf{18.20} & \textbf{18.58} & \textbf{24.78} & \underline{15.49} & \textbf{2.44} \\
\midrule
$\Delta S$ vs.\ best baseline & $+6.84$ & $+1.74$ & $+5.16$ & $+24.11$ & $+8.80$ & $+12.93$ & & \\
\bottomrule
\end{tabular*}
\caption{Acquisition--retention score $S=A-F$ (pp; $M_0$ has $S=0$), with six-setting means of acquisition $A$ and forgetting $F$. In each column, bold marks the best value and underlining the second best (highest $S$ or $A$, lowest $F$); in the $S$ columns the underlined entry is therefore the strongest baseline. The shaded row is InstructReplay. Figure~\ref{fig:af} shows the per-setting $A$/$F$ decomposition; Tables~\ref{tab:cc-full}--\ref{tab:pmc-full} give all raw scores.}
\label{tab:results-1}
\end{table*}

\begin{table*}[t]
\centering\small
\setlength{\tabcolsep}{1.6pt}
\begin{tabular*}{\textwidth}{@{\extracolsep{\fill}}llrrrrrrrrr@{}}
\toprule
 & & \multicolumn{5}{c}{Mean change vs.\ $M_0$ (pp)} & Length & \multicolumn{3}{c}{Mean of six (pp)} \\
\cmidrule(lr){3-7}\cmidrule(lr){9-11}
Replay context (method) & Target & IFEval & MATH & BBH & MMLU & HE & IFEval & $A\uparrow$ & $F\downarrow$ & $S\uparrow$ \\
\midrule
None (Naive SFT) & -- & $-44.9$ & $-28.0$ & $-36.0$ & \underline{$-0.4$} & $-30.5$ & 0.11 & 15.2 & 28.1 & $-12.8$ \\
Domain-answer prefixes (Domain-KL)$^\ast$ & soft & $-43.3$ & $-12.7$ & $-23.7$ & $\mathbf{+0.1}$ & $-21.4$ & 0.19 & 14.0 & 20.6 & $-6.5$ \\
BOS samples, no template (BOS replay) & soft & $-14.2$ & $-6.2$ & $-6.2$ & $-0.8$ & $-15.9$ & 0.62 & \underline{15.6} & 9.2 & 6.4 \\
Instruction prompts, dataset answers (Gold replay) & dataset & $-7.4$ & $-4.3$ & $-6.7$ & $-5.3$ & $-10.9$ & 0.74 & \textbf{15.7} & 7.5 & 8.2 \\
Instruction responses (Hard) & sampled & \underline{$-4.0$} & \underline{$-0.3$} & $\mathbf{+0.8}$ & $-0.9$ & \underline{$-5.7$} & \underline{0.82} & 15.0 & \underline{3.1} & \underline{11.9} \\
\fullrowcolor{oursrow}{\textwidth}Instruction responses (InstructReplay) & soft & $\mathbf{-2.6}$ & $\mathbf{-0.2}$ & \underline{$-0.7$} & $-1.3$ & $\mathbf{-1.0}$ & \textbf{0.85} & 15.5 & \textbf{2.4} & \textbf{13.1} \\
\bottomrule
\end{tabular*}
\caption{Moving the retention constraint, averaged over six settings. All replay variants use $\lambda=0.25$, a 1:1 sequence ratio and 300 steps; BOS replay and Hard also use exactly InstructReplay's response-token budget, so each differs from it mainly in the replay context or the target; Gold replay differs from Hard mainly in training on the dataset's answers instead of $M_0$'s (2.8M response tokens per model). HE is HumanEval+. Length is the median IFEval response length relative to $M_0$. In each column, bold marks the best value and underlining the second best (smallest decline, length closest to $M_0$, highest $A$ and $S$, lowest $F$); the shaded row is InstructReplay. $^\ast$Domain-KL constrains about one tenth as many positions (0.20--0.36M versus 3.0--3.5M per model). Per-setting scores are in Table~\ref{tab:results-18}.}
\label{tab:where}
\end{table*}

\begin{figure*}[t]
\centering
\begin{tikzpicture}
\begin{groupplot}[group style={group size=3 by 2, horizontal sep=0.9cm, vertical sep=0.55cm, xlabels at=edge bottom, xticklabels at=edge bottom},
  width=0.33\textwidth, height=3.9cm, xmin=-2, xmax=46, axis lines*=left,
  ticklabel style={font=\scriptsize}, title style={font=\small\bfseries, yshift=-3pt},
  xlabel={Forgetting $F$ (pp) $\downarrow$}, xlabel style={font=\small},
  ymajorgrids, xmajorgrids, grid style={gray!15}, method classes, clip=false]
\nextgroupplot[ymin=-3, ymax=14, title={SmolLM3-3B}, ylabel={\textbf{CommonCorpus}\\[-1pt]$A$ (pp) $\uparrow$}, ylabel style={font=\small, align=center}]
\addplot[fill=ircol!7, draw=none, forget plot] coordinates {(0,0) (14,14) (0,14)} -- cycle;
\addplot[dashed, gray, forget plot, domain=0:14] {x};
\addplot[scatter, only marks, scatter src=explicit symbolic] table[x=F, y=A, meta=cls] {figures/data/af-commoncorpus-smollm3.dat};
\nextgroupplot[ymin=-3, ymax=14, title={Qwen3-4B}]
\addplot[fill=ircol!7, draw=none, forget plot] coordinates {(0,0) (14,14) (0,14)} -- cycle;
\addplot[dashed, gray, forget plot, domain=0:14] {x};
\addplot[scatter, only marks, scatter src=explicit symbolic] table[x=F, y=A, meta=cls] {figures/data/af-commoncorpus-qwen3_4b.dat};
\nextgroupplot[ymin=-3, ymax=14, title={Phi-4-mini}]
\addplot[fill=ircol!7, draw=none, forget plot] coordinates {(0,0) (14,14) (0,14)} -- cycle;
\addplot[dashed, gray, forget plot, domain=0:14] {x};
\addplot[scatter, only marks, scatter src=explicit symbolic] table[x=F, y=A, meta=cls] {figures/data/af-commoncorpus-phi4_mini.dat};
\nextgroupplot[ymin=-2, ymax=31, ylabel={\textbf{PMC}\\[-1pt]$A$ (pp) $\uparrow$}, ylabel style={font=\small, align=center}, legend to name=aflegend, legend columns=6, legend cell align=left, legend style={font=\scriptsize, draw=none, /tikz/every even column/.append style={column sep=5pt}}]
\addplot[fill=ircol!7, draw=none, forget plot] coordinates {(0,0) (31,31) (0,31)} -- cycle;
\addplot[dashed, gray, forget plot, domain=0:31] {x};
\addplot[scatter, only marks, scatter src=explicit symbolic] table[x=F, y=A, meta=cls] {figures/data/af-pmc-smollm3.dat};
\legend{$M_0$, Naive SFT, LoRA-SFT, Self-Rewrite, SDFT, DFT, RAFT, Domain-KL, BOS replay, Hard, InstructReplay}
\nextgroupplot[ymin=-2, ymax=31]
\addplot[fill=ircol!7, draw=none, forget plot] coordinates {(0,0) (31,31) (0,31)} -- cycle;
\addplot[dashed, gray, forget plot, domain=0:31] {x};
\addplot[scatter, only marks, scatter src=explicit symbolic] table[x=F, y=A, meta=cls] {figures/data/af-pmc-qwen3_4b.dat};
\nextgroupplot[ymin=-2, ymax=31]
\addplot[fill=ircol!7, draw=none, forget plot] coordinates {(0,0) (31,31) (0,31)} -- cycle;
\addplot[dashed, gray, forget plot, domain=0:31] {x};
\addplot[scatter, only marks, scatter src=explicit symbolic] table[x=F, y=A, meta=cls] {figures/data/af-pmc-phi4_mini.dat};
\end{groupplot}
\end{tikzpicture}

\vspace{1pt}\ref{aflegend}
\caption{Acquisition versus forgetting for every method, including the two replay controls (open markers). Upper left is better; the shaded region has $S=A-F>0$ and the dashed line marks $S=0$. Forgetting shares one scale; acquisition scales differ by domain.}
\label{fig:af}
\end{figure*}
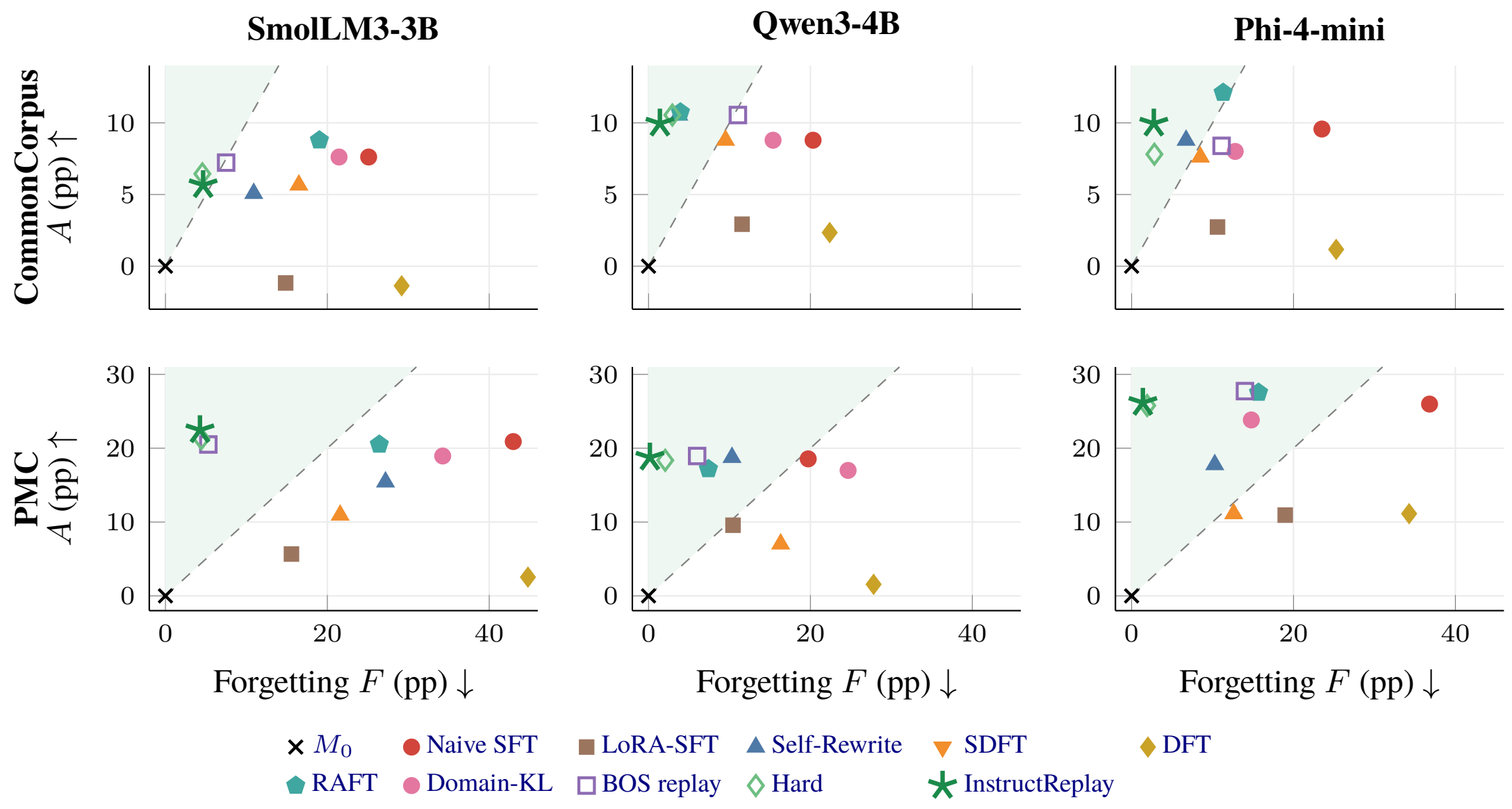

\textbf{Instruction-context replay removes most forgetting and keeps acquisition on par with standard fine-tuning.} InstructReplay has the highest $S$ in all six settings (Table~\ref{tab:results-1}). The gain comes almost entirely from reduced forgetting. Averaged over the six settings, InstructReplay's general decline is 2.44 points against 28.07 for Naive SFT, and it stays below 4.7 in every setting, whereas the lowest-forgetting baseline in each setting loses at least 3.73. No general benchmark loses more than 3 points on average under InstructReplay, whereas standard fine-tuning loses 28 or more on each generative benchmark. Its acquisition is within 1.95 points of Naive SFT or higher in every setting, and its average of 15.49 is second only to RAFT's 16.15. InstructReplay Pareto-dominates 35 of the 42 baseline points in the $A$--$F$ plane (Figure~\ref{fig:af}); each exception acquires at most 3.1 points more but forgets at least 2.33 points more.

The closest competitors are those that also keep the model near its own distribution. Self-Rewrite, which trains on $M_0$'s rewrites of the gold answers, is the strongest baseline on CommonCorpus, and RAFT, which adds distillation to refined answers, is the strongest on PMC. The closest case is CommonCorpus Qwen3, where InstructReplay acquires 0.59 points less than Self-Rewrite but forgets 2.33 points less, a margin of 1.74 in $S$. Elsewhere the margin is at least 5.16 points and reaches 24.11, driven by forgetting: on PMC SmolLM3, for example, RAFT forgets 26.42 points against InstructReplay's 4.26.

\textbf{InstructReplay keeps the initial model's answer length.} Its median response length stays within 15\% of $M_0$'s on IFEval, MATH-500 and BBH, whereas standard fine-tuning at least halves it (Appendix~\ref{app:length}). On questions that $M_0$ answers correctly, 69\% of InstructReplay's responses fall in the near-$M_0$ band of Figure~\ref{fig:length-main}b, where failures are rarest, against 24\% for standard fine-tuning.

\subsection{Where to constrain and what to match}\label{sec:where}

\textbf{The location of the constraint drives retention.} Table~\ref{tab:where} holds the objective fixed and moves it. Placing the KL on domain-answer prefixes barely helps: IFEval still falls by 43 points and responses remain a fifth of their original length. Placing it on BOS-conditioned samples, which are generated without the chat template, recovers about two thirds of the IFEval loss. Placing it on the model's own answers to general instructions recovers almost all of it, and restores response length to 85\% of $M_0$'s. Acquisition stays between 14.0 and 15.7 points throughout, so the differences in $S$ come from forgetting. At an identical response-token budget, InstructReplay has higher $S$ than BOS replay in all six settings, by 1.28 to 11.05 points, and lower $F$ in each (Appendix~\ref{app:replay}).

\textbf{Self-generated answers reduce average forgetting.} Gold replay uses the prompts and loss of Hard but trains on the collection's own answers. It acquires as much as InstructReplay, 15.7 against 15.5 points on average, and exceeds the strongest of the seven baselines in five of six settings, yet it forgets more than twice as much as Hard, which replays $M_0$'s answers with the same loss, 7.5 against 3.1 points. Its $S$ is lower than InstructReplay's in all six settings, by at least 2.06 points. On SmolLM3 it lowers MMLU by 18.2 and 12.3 points, where no other method loses more than 4.6 (Appendix~\ref{app:replay}). With the instruction prompts held fixed, self-generated answers preserve general performance better on average than the dataset answers tested here.

\textbf{Soft and sampled targets perform similarly at the shared coefficient.} Replacing the soft targets with the saved tokens themselves, as Hard does, leaves forgetting almost unchanged: 3.1 versus 2.4 points on average. InstructReplay has the higher $S$ in five settings and Hard in one, CommonCorpus SmolLM3, by 0.88 points; no difference exceeds 2.29 points. At a fixed prefix, the expected sampled-token gradient equals the forward-KL gradient (Appendix~\ref{app:loss}); Hard can omit the reference forward pass from the replay loss. At this shared coefficient, both preserve assistant behavior, while replay context and answer source produce larger differences in average forgetting.

\subsection{How robust are the conclusions?}\label{sec:robust}

\begin{table}[t]
\centering\small
\setlength{\tabcolsep}{4pt}
\begin{tabular}{@{}llrr@{}}
\toprule
Setting & vs. & $\Delta S$ & 95\% CI \\
\midrule
CC Smol & S-R & 6.84 & [$-0.40$, 14.11] \\
CC Qwen & S-R & 1.74 & [$-5.20$, 8.36] \\
CC Phi & S-R & 5.16 & [$-2.15$, 13.01] \\
PMC Smol & RAFT & 24.11 & [14.28, 32.91] \\
PMC Qwen & RAFT & 8.80 & [1.33, 16.32] \\
PMC Phi & RAFT & 12.93 & [3.58, 21.22] \\
\bottomrule
\end{tabular}
\caption{InstructReplay minus the strongest baseline (S-R: Self-Rewrite; CC: CommonCorpus). The CI is a paired cluster-bootstrap percentile interval, Bonferroni-corrected over all 36 pairs among the eight methods and $M_0$ in each setting.}
\label{tab:robust}
\end{table}

\textbf{Question sampling.} Against the strongest baseline, the corrected interval excludes zero in the three PMC settings (Table~\ref{tab:robust}), and across all 42 InstructReplay--baseline pairs it excludes zero in 37 (Appendix~\ref{app:stats}). Appendix Table~\ref{tab:control-ci} gives the corresponding intervals against the three replay controls.

\textbf{Training runs.} Repeating training three times for InstructReplay, Self-Rewrite and RAFT leaves the ranking unchanged. InstructReplay has the highest mean $S$ in every setting and exceeds both baselines in 35 of the 36 run-by-run comparisons (Appendix~\ref{app:runs}).

\textbf{Weighting of acquisition and forgetting.} $S$ weights acquisition and forgetting equally. Under $S_\beta=A-\beta F$, InstructReplay has the highest score among the seven baselines in every setting whenever $\beta>0.31$; the binding case is RAFT on CommonCorpus Qwen3, and on PMC SmolLM3 and Qwen3 it is highest for every $\beta>0$. If gains on some benchmarks are instead allowed to offset losses on others, $U=A+\frac15\sum_k(G_k-G_k^0)$, InstructReplay still has the highest value in every setting.

\textbf{Answer quality and judging.} Requiring every designated detail and no unsupported material claims, which we call full-answer confirmation, leaves InstructReplay ahead of all seven baselines, BOS replay and Hard in all six settings, with acquisition between 6.84 and 31.64 points. Scoring CommonCorpus over its 374 eligible questions only rescales acquisition and leaves InstructReplay ahead of every baseline and BOS replay. On CommonCorpus, BOS replay and Hard each leave 128 questions without a judge majority, against 30 for InstructReplay; Appendix~\ref{app:scoring} bounds the effect of these labels.

\section{Analysis}\label{sec:analysis}

\textbf{Replay source.} The diagnosis in \S\ref{sec:diagnosis} ties the loss of general ability to responses that drift away from $M_0$'s. Replay counters this drift best when its targets are $M_0$'s own responses in assistant turns. Domain-answer prefixes cover only short factual answers, and BOS samples lie outside the chat template. The dataset's answers are assistant turns in a different style, so fitting them moves the model away from $M_0$ in another direction (Appendix~\ref{app:replay}). Gold replay keeps $M_0$'s MATH-500 response length yet loses 4.3 points there, against 0.2 for InstructReplay, so matching length alone does not preserve accuracy. Updates driven by a model's own outputs have likewise been found to forget less than training on external targets \citep{shenfeld2025razor,chen2026retaining}.

\textbf{Knowledge-level changes.} InstructReplay keeps assistant behavior close to $M_0$'s, and its MMLU score changes little, by $-1.35$ points on average. The largest change, $-4.56$ on CommonCorpus SmolLM3, does not come from output format, since the four options keep nearly all of the probability mass at the scored position, as they do for $M_0$. It also varies across training runs (Appendix~\ref{app:runs}).

\textbf{Retention weight.} We chose $\lambda=0.25$ among $\{0, 0.25, 1, 4\}$ on CommonCorpus with SmolLM3 and Qwen3, where it gave the highest $S$ for both (Appendix~\ref{app:lambda}). Relative to no replay, which at $\lambda=0$ is identical to Naive SFT, the selected replay weight lowers $F$ from 25.10 to 4.66 for SmolLM3 and from 20.33 to 1.40 for Qwen3. Larger weights trade acquisition for retention unevenly: at $\lambda=4$, Qwen3 forgets slightly less but loses a quarter of its acquisition. Because $\lambda$ was selected on these two settings, the four transfer settings, Phi-4-mini and all of PMC, are the fairer test of the recipe, and there InstructReplay leads the strongest baseline by 5.16 points or more in $S$.

\textbf{Cost.} InstructReplay doubles the sequences per step and replays about 3M response tokens per model, roughly an order of magnitude more than its domain tokens. Yet its PMC training takes at most 2.22 GPU-hours per model, against at least 4.83 for RAFT, and the buffer is generated once per model and shared across domains (Appendix~\ref{app:lambda}).

\section{Conclusion}\label{sec:conclusion}

Fine-tuning small instruction-tuned models on new facts mainly damaged their generative behavior: generative benchmarks collapsed and responses shrank, while likelihood-scored MMLU barely moved. Replaying the initial model on general instructions largely prevented this damage at little cost to acquisition. In the controlled comparisons, assistant-turn replay gave lower average forgetting than BOS-conditioned and domain-answer replay, and self-generated answers gave lower average forgetting than the tested dataset answers. Soft and sampled targets performed similarly at the shared coefficient. For practitioners, about ten thousand general instructions answered once by the model being adapted are an inexpensive safeguard. Small MMLU declines under replay and larger models remain open.

\section*{Limitations}

\textbf{Evaluation.} Domain acquisition relies on model judges without human calibration, judges of the same families also score RAFT's training targets, and invalid-vote rates differ between methods. Some evaluation questions cue part of the answer. Publication dates alone cannot rule out that related facts appeared in pretraining.

\textbf{Coverage of general ability.} Our evaluation covers five general benchmarks that overlap in kind with the replay mixture; whether retention extends to abilities absent from the replay data, such as safety behavior or long-context use, remains untested. MMLU is scored by likelihood and is therefore insensitive to generation failures.

\textbf{Comparisons.} The replay controls use the shared settings in \S\ref{sec:setup-baselines}; configuration differences and the Gold replay formatting diagnostic are detailed in Appendix~\ref{app:replay}. Method-specific tuning remains open. The study covers three models from different families, all at the 4B scale, two domains and short training (300 steps), and the six settings share general-benchmark questions and are therefore not independent.

\section*{Ethics and Reproducibility}

Biomedical scores measure factual answering about research articles and do not support clinical use. PMC sources were selected for CC BY 4.0 licenses; the UltraData repository declares Apache-2.0 and requires compliance with incorporated upstream licenses, and full-text redistribution awaits row-level clearance. The accompanying supplement provides the paper source, result tables, hash-anonymized domain labels, scoring materials and a verification script (Appendix~\ref{app:repro}).

\bibliography{references}

\clearpage
\appendix
\raggedbottom
\setcounter{dbltopnumber}{4}
\renewcommand{\dbltopfraction}{0.95}
\renewcommand{\textfraction}{0.05}
\renewcommand{\dblfloatpagefraction}{0.85}
\renewcommand{\floatpagefraction}{0.85}
\renewcommand{\topfraction}{0.95}
\setcounter{topnumber}{4}
\setcounter{totalnumber}{6}
\makeatletter
\setlength{\@fptop}{0pt}
\setlength{\@fpsep}{16pt}
\setlength{\@fpbot}{0pt plus 1fil}
\setlength{\@dblfptop}{0pt}
\setlength{\@dblfpsep}{16pt}
\setlength{\@dblfpbot}{0pt plus 1fil}
\makeatother
\section{Properties of the Retention
Loss}\label{app:loss}

For a fixed prefix, let \(a\) be student logits,
\(q=\operatorname{softmax}(a)\), and \(q_0\) the frozen reference
distribution. Expanding forward KL gives
\(\ell(a)=\sum_v q_{0v}\log q_{0v}-\sum_v q_{0v}\log q_v\).
Differentiation yields \(\nabla_a\ell=q-q_0\) and
\(\nabla_a^2\ell=\operatorname{diag}(q)-qq^\top\). At \(a=a_0\), both
the loss and its first derivative vanish. Its Hessian is positive
semidefinite, since for any vector \(u\),

\[
u^\top(\operatorname{diag}(q_0)-q_0q_0^\top)u
 =\operatorname{Var}_{v\sim q_0}(u_v)\geq 0.
\]

For smooth logits \(a_\theta(h)\), the parameter-space Hessian at
\(\theta_0\) is \(J_h^\top(\operatorname{diag}(q_0)-q_0q_0^\top)J_h\).
The terms involving second derivatives of the logits vanish because the
logit gradient is zero at that point. Choosing a sequence uniformly from
\(\mathcal B_R\), then a position uniformly from \(T(z)\), defines

\[
\widehat\mu_R=\frac{1}{N_R}\sum_{(r,z)\in\mathcal B_R}\frac{1}{|T(z)|}\sum_{t\in T(z)}\delta_{(r,z_{<t})}.
\]

Averaging the fixed-prefix Hessian gives

\[
\begin{aligned}
H_R&=\mathbb E_{h\sim\widehat\mu_R}[J_h^\top K_hJ_h],\\
K_h&=\operatorname{diag}(q_0^h)-q_0^h(q_0^h)^\top,\\
\mathcal L_R(\theta_0+\delta)&=\tfrac12\delta^\top H_R\delta+o(\|\delta\|^2).
\end{aligned}
\]

Here \(J_h\) is the logit Jacobian at \(\theta_0\). The expansion
assumes finite logits, twice differentiability near \(\theta_0\), and a
fixed mask and prefix set. The positive-semidefinite Hessian describes
local sensitivity on the saved replay contexts and may have a null
space.

For a hard token \(Z\sim q_0\) at a fixed prefix, the logit gradient at
initialization is \(q_0-e_Z\). Its expectation is zero and its
covariance is \(\operatorname{diag}(q_0)-q_0q_0^\top\), with expected
squared norm \(1-\|q_0\|_2^2\). At any student distribution \(q\), the
sampled-token gradient \(q-e_Z\) has expectation \(q-q_0\), the
forward-KL gradient, so cross-entropy on tokens sampled from \(q_0\) is an
unbiased single-sample estimate of the same objective up to the constant
entropy of \(q_0\). These identities condition on the prefix
and sampling distribution. Capping responses and averaging each sequence
by its valid length defines the empirical prefix weighting
\(\widehat\mu_R\) above, distinct from the unnormalized KL between
complete response distributions.

For a retention-only update at the reference point, zero parameter
change also requires zero weight decay, fresh optimizer state, and
identical deterministic student/reference behavior. In mixed training,
the domain gradient can still update the student. Appendix~\ref{app:training} summarizes
implementation checks.

\section{Data and Evaluation
Protocol}\label{app:data}

\subsection{Domain data
construction}\label{b.1.-domain-data-construction}

CommonCorpus \citep{pleias2024commoncorpus} energy/engineering passages
are organized into source-supported fact groups. A 9,600-row training
set and 512 evaluation questions ask about the same taught facts using
different formulations. The PMC Open Access Subset \citep{ncbi2026pmc}
supplies 4,500 fixed article candidates: license/date checks retain
2,161 and topic screening retains 1,661. Eligible articles declare CC BY
4.0, online publication from 1 May through 31 August 2026, and
acceptance in 2026 or later. Fixed hash selection retains 3,561 of the 4,170
qualified fact groups, the same number as in CommonCorpus. The training questions cover 1,450 articles; the evaluation questions cover 512 taught facts from 440 articles. PMC admission requires M0's confirmation rate to stay below 30\% for each model. The criterion applies to the whole collection and does not select questions from model-specific errors.

For both datasets, Qwen3.5-35B proposes alternate questions for existing
answer propositions; Qwen3.8-27B audits evidence support, including a
negative control in each request. Generation uses temperature 0, top-p
1, and a 512-token limit; audit responses allow 1,024 tokens. The
alternate-question generation stage receives an existing training
question, its answer proposition and source evidence, without the evaluation question wording. The builder accepts only candidates that
pass its lexical checks and all seven checks: evidence answerability,
question--answer relevance, answer support, self-contained wording,
natural wording, no answer leakage, and meaningful rewording;
abstentions are excluded. The supplement's data cards give the retained counts at each stage.

The supplement's confirmation-transition table separates previously
unconfirmed answers that become confirmed from confirmed answers that
lose confirmation. For CommonCorpus Smol InstructReplay, these counts
are 59 and 30, yielding net acquisition \(100(59-30)/512=5.66\)
pp.~Unconfirmed answers combine valid negative judgments and unresolved
cases. For each of the 54 Table~\ref{tab:results-1} rows, the net
transition equals \(C-C^0\); the supplement provides all transition counts.

\begin{table*}[!tbp]
\centering
\small
\setlength{\tabcolsep}{3pt}
\begin{tabularx}{\textwidth}{@{}>{\raggedright\arraybackslash}X>{\raggedright\arraybackslash}X>{\raggedright\arraybackslash}X>{\raggedright\arraybackslash}X@{}}
\toprule
Domain & Train/eval questions & Train/eval documents & Train/eval fact groups \\
\midrule
CommonCorpus & 9,600 / 512 & 961 / 386 & 3,561 / 512 \\
PMC & 9,600 / 512 & 1,450 / 440 & 3,561 / 512 \\
\bottomrule
\end{tabularx}
\caption{Dataset sizes. Both domains assess taught facts with rephrased evaluation questions. Source identifiers and construction metadata are in the supplement's data cards.}
\label{tab:results-4}
\end{table*}

For PMC, the initial Smol/Qwen/Phi checkpoints confirm 83/95/57 of 512
answers. The training sources cover genetics/genomics, immunology,
neuroscience, oncology and infection research.

The CommonCorpus split is defined by a training manifest with file hashes: 6,889 source tasks define 3,564
source fact groups; normalization removes 213 duplicate tasks and
retains 3,561 distinct fact groups. A fact with multiple question
formulations supplies a held-out formulation while remaining questions
expose that fact in training; singleton facts stay training-only. The
9,600-row pack contains 6,164 retained questions and 3,436 accepted
augmentations. The passage filter uses language and engineering keyword profiles; passages are thus selected by vocabulary
rather than by subject classification, and the collection includes
technical computing articles such as the first example below. Parent records retain source identifiers and license labels.
PMC uses the article selection described above; 6,610 retained
questions plus 2,990 augmentations form its training set.

\textbf{Source-linked examples.}

A CommonCorpus fact comes from the SPARF article of
\citet{hamdi2023sparf}, released under CC BY 4.0. The source states:
``INGP {[}46{]} uses multi-resolution voxel hashing to perform a
real-time rendering of radiance fields.'' Its scored CORE proposition
links multi-resolution voxel hashing to real-time rendering. The evaluation question supplies the mechanism but leaves the real-time
effect to be stated; the reference answer also discusses MLP redundancy,
outside the scored CORE. This makes the question a test of the effect
given a mechanism cue. The training question is ``How does INGP perform
real-time rendering of radiance fields?'' The evaluation question is
``How does INGP use multi-resolution voxel hashing to address rendering
limitations in radiance fields?'' The supplement provides the fact card and both question--answer pairs. The public article contains the quoted passage.

The following case comes from the PMC evaluation. The
simvastatin study of \citet{li2026simvastatin} was released under
Creative Commons Attribution. Its abstract states: ``In vivo, pmPLGA@SV
treatment dramatically reduced the cerebral infarct volume, exhibiting
significant superiority over free SV.'' The fact card makes reduction
relative to free simvastatin its single CORE proposition. The training
question asks ``What was the observed effect of pmPLGA@SV on cerebral
infarct volume relative to free simvastatin in vivo?'' The separate evaluation question asks how treatment affected that volume compared
with free simvastatin in the in vivo model.

A fixed hash-ranked sample of 16 evaluation questions per domain was inspected without using model outputs. Against the
scored CORE, one CommonCorpus question directly supplied part of the
target relation, 13 did not, and two were uncertain (one missing a CORE
card). The 16 PMC questions did not directly supply their target
propositions. Entity or topic overlap alone was not counted as a cue.
The separately illustrated INGP example is outside this sample. This
small sample illustrates question cues.

\subsection{Replay population}\label{b.2.-replay-population}

The source is \mbox{\texttt{openbmb/UltraData-SFT-2605}}, revision
\nolinkurl{affda6aca75e7cff78e73f93ad08d4c3b01f097c}. The 9,600 prompts
allocate 1,920 each to instruction following, academic tasks,
mathematics, coding, and translation. Mathematics splits equally between
Math and Multi-lang-Math. Selection uses deterministic hash ranking and
normalized prompt deduplication, preserving the parent train/monitor
split, and excludes prompts over 4,096 UTF-8 bytes. Selection is
independent of model scores. Prompt lengths range from 27 to 4,078 bytes, with a mean of 779.

Final assistant turns are removed, and the remaining conversation must
end with a user message. Earlier system/user/assistant turns remain as
input context; 25 of 9,600 prompts contain earlier assistant turns. Each
model generates its own targets once, without correctness filtering. A
lexical comparison against the General evaluation questions and few-shot
examples found no normalized exact, substring, or trigram candidates.
The check covers lexical overlap; semantic overlap and pretraining
exposure remain unmeasured.

\subsection{Replay generation and
masks}\label{b.3.-replay-generation-and-masks}

Native chat templates use \texttt{enable\_thinking=False}.
General-response sampling uses temperature 1, top-p 1, top-k 0, and at
most 512 new tokens; generation stops at the model's EOS and includes
that terminal token in supervision. Phi accepts its two native stopping
IDs. Prompt and padding positions are masked; nonempty response tokens
are retained without a separate reasoning-token mask. Materialization
rejects rows exceeding the training length limit. In Smol/Qwen/Phi
order, response lengths average 363.11/351.59/312.14 tokens, with
medians 424/417/329 and 90th/95th percentiles of 512 for all three.
Actual length-cap stops occur in 3,960/3,975/3,251 of 9,600 responses;
EOS at exactly 512 tokens accounts for the difference from raw length
counts. All retained responses contain at least one target token.

\subsection{Domain scoring}\label{app:scoring}

\textbf{Eligibility and timing.} The CommonCorpus rule requires a
mechanically valid fact card and usable-card judgments from both Gemma
and GPT-OSS. Mechanical validity requires answerability, nonempty facts
and requirements, unique fact IDs, at least one CORE fact, and verbatim
source quotes. Of 512 records, 68 are marked unanswerable by the card
builder, four have no resolved builder result, and 26 fail quote
matching. Among the 414 mechanically valid cards, four fail both audits,
24 fail only Gemma's audit and 12 fail only GPT-OSS's audit, leaving 374
eligible. These mutually exclusive counts reconstruct the fixed mask
distributed with the supplement.

We keep all 512 questions in the denominator, so the maximum attainable CommonCorpus rate is \(374/512=73.05\%\).
Eligibility uses the question, source evidence, reference answer and fact-card audits, without the candidate answers.

A fact card identifies each required proposition, its CORE or
REQUIRED\_DETAIL role, and an exact source quote. Scoring prompts
contain the question, source evidence, key, and one candidate answer,
without a method name or competing answer. Judges assign SUPPORTED,
OMITTED, CONTRADICTED, or UNCLEAR per fact and separately label material
extra claims. Valid records must cover the exact fact IDs and cite
matching answer/source substrings. All CORE facts must be supported; a
relevant contradiction makes the core vote negative. Unsupported
additional claims are reported separately and do not automatically veto
core confirmation; they do affect the stricter full-answer measure.
Ineligible tasks, unresolved records, and insufficient valid votes
remain unconfirmed in the denominator of 512. Two positive valid core
votes confirm an answer.

Judging uses temperature 0, top-p 1, and at most two total format
attempts per call, without hidden client retries. In the focused CC
records, Smol uses a 180-second timeout and Qwen/Phi use 600 seconds;
judge token budgets range from 3,072 to 8,192, as recorded per run. Qwen
and Gemma disable thinking; GPT-OSS uses low reasoning effort. The supplement gives the complete prompts and the quote-validation function. Constructed
positive/negative controls check implementation behavior. Prompt-order
and repeat diagnostics are recorded separately from the scoring votes.

\textbf{Full-answer sensitivity.} Core confirmation asks whether the
taught core fact is stated correctly; full confirmation additionally
requires designated details and no material unsupported extras. The
core/full gap combines omitted details, unsupported extras and
unresolved judgments. Table~\ref{tab:results-5} compares methods using each measure's own M0 count and the same F.

\begin{table*}[!tbp]
\centering
\small
\setlength{\tabcolsep}{3pt}
\begin{tabular*}{\textwidth}{@{\extracolsep{\fill}}lrrrrrr@{}}
\toprule
Method & Core n & Full n & \(A_{\mathrm{core}}\) & \(A_{\mathrm{full}}\) & \(F\) & \(S_{\mathrm{full}}\) \\
\midrule
\multicolumn{7}{l}{\textit{CommonCorpus --- SmolLM3-3B}} \\
\addlinespace[2pt]
M0 & 50 & 4 & 0.00 & 0.00 & 0.00 & 0.00 \\
Naive SFT & \underline{89} & \textbf{57} & \underline{7.62} & \textbf{10.35} & 25.10 & -14.75 \\
Self-Rewrite & 76 & 14 & 5.08 & 1.95 & 10.91 & -8.96 \\
RAFT & \textbf{95} & 22 & \textbf{8.79} & 3.52 & 19.02 & -15.50 \\
BOS replay & 87 & \textbf{57} & 7.23 & \textbf{10.35} & \underline{7.50} & \underline{2.85} \\
\fullrowcolor{oursrow}{\textwidth}InstructReplay & 79 & \underline{53} & 5.66 & \underline{9.57} & \textbf{4.66} & \textbf{4.91} \\
\midrule
\multicolumn{7}{l}{\textit{CommonCorpus --- Qwen3-4B}} \\
\addlinespace[2pt]
M0 & 38 & 6 & 0.00 & 0.00 & 0.00 & 0.00 \\
Naive SFT & 83 & 47 & 8.79 & 8.01 & 20.33 & -12.32 \\
Self-Rewrite & \underline{92} & 35 & \underline{10.55} & 5.66 & \underline{3.73} & 1.93 \\
SDFT & 83 & 30 & 8.79 & 4.69 & 9.51 & -4.82 \\
RAFT & \textbf{93} & 46 & \textbf{10.74} & 7.81 & 3.94 & \underline{3.87} \\
BOS replay & \underline{92} & \textbf{57} & \underline{10.55} & \textbf{9.96} & 11.04 & -1.08 \\
\fullrowcolor{oursrow}{\textwidth}InstructReplay & 89 & \underline{49} & 9.96 & \underline{8.40} & \textbf{1.40} & \textbf{7.00} \\
\midrule
\multicolumn{7}{l}{\textit{CommonCorpus --- Phi-4-mini}} \\
\addlinespace[2pt]
M0 & 33 & 9 & 0.00 & 0.00 & 0.00 & 0.00 \\
Naive SFT & 82 & \textbf{49} & 9.57 & \textbf{7.81} & 23.51 & -15.70 \\
Self-Rewrite & 78 & 23 & 8.79 & 2.73 & \underline{6.73} & -4.00 \\
SDFT & 72 & 36 & 7.62 & 5.27 & 8.50 & \underline{-3.23} \\
RAFT & \textbf{95} & 35 & \textbf{12.11} & 5.08 & 11.31 & -6.23 \\
BOS replay & 76 & \underline{44} & 8.40 & \underline{6.84} & 11.12 & -4.28 \\
\fullrowcolor{oursrow}{\textwidth}InstructReplay & \underline{84} & \underline{44} & \underline{9.96} & \underline{6.84} & \textbf{2.74} & \textbf{4.10} \\
\midrule
\multicolumn{7}{l}{\textit{PMC --- SmolLM3-3B}} \\
\addlinespace[2pt]
M0 & 83 & 23 & 0.00 & 0.00 & 0.00 & 0.00 \\
Naive SFT & \underline{190} & \underline{179} & \underline{20.90} & \underline{30.47} & 42.97 & -12.50 \\
LoRA-SFT & 112 & 105 & 5.66 & 16.02 & 15.57 & 0.45 \\
Self-Rewrite & 162 & 115 & 15.43 & 17.97 & 27.19 & -9.22 \\
RAFT & 188 & 124 & 20.51 & 19.73 & 26.42 & -6.69 \\
BOS replay & 188 & 175 & 20.51 & 29.69 & \underline{5.30} & \underline{24.39} \\
\fullrowcolor{oursrow}{\textwidth}InstructReplay & \textbf{198} & \textbf{185} & \textbf{22.46} & \textbf{31.64} & \textbf{4.26} & \textbf{27.38} \\
\midrule
\multicolumn{7}{l}{\textit{PMC --- Qwen3-4B}} \\
\addlinespace[2pt]
M0 & 95 & 44 & 0.00 & 0.00 & 0.00 & 0.00 \\
Naive SFT & 190 & \underline{175} & 18.55 & \underline{25.59} & 19.73 & 5.86 \\
Self-Rewrite & \underline{191} & 157 & \underline{18.75} & 22.07 & 10.31 & 11.76 \\
RAFT & 183 & 167 & 17.19 & 24.02 & 7.40 & 16.62 \\
BOS replay & \textbf{192} & \textbf{179} & \textbf{18.95} & \textbf{26.37} & \underline{6.00} & \underline{20.37} \\
\fullrowcolor{oursrow}{\textwidth}InstructReplay & \underline{191} & 172 & \underline{18.75} & 25.00 & \textbf{0.17} & \textbf{24.83} \\
\midrule
\multicolumn{7}{l}{\textit{PMC --- Phi-4-mini}} \\
\addlinespace[2pt]
M0 & 57 & 32 & 0.00 & 0.00 & 0.00 & 0.00 \\
Naive SFT & 190 & 174 & 25.98 & 27.73 & 36.81 & -9.08 \\
Self-Rewrite & 148 & 125 & 17.77 & 18.16 & \underline{10.27} & 7.89 \\
RAFT & \underline{198} & \textbf{186} & \underline{27.54} & \textbf{30.08} & 15.68 & 14.40 \\
BOS replay & \textbf{199} & \underline{184} & \textbf{27.73} & \underline{29.69} & 14.00 & \underline{15.69} \\
\fullrowcolor{oursrow}{\textwidth}InstructReplay & 191 & 179 & 26.17 & 28.71 & \textbf{1.39} & \textbf{27.32} \\
\bottomrule
\end{tabular*}
\caption{Core/full counts and acquisition--retention sensitivity (denominator 512). \(A_{\mathrm{full}}=100(C_{\mathrm{full}}-C^0_{\mathrm{full}})/512\) and \(S_{\mathrm{full}}=A_{\mathrm{full}}-F\); all A/F/S entries are pp.~Core and full use their respective M0 counts.  Shaded rows mark InstructReplay. Within each setting, bold/underline mark the best/second-best shown adapted result (higher counts, $A$ and $S$, lower $F$); M0 is a reference. Ties share marks.}
\label{tab:results-5}
\end{table*}

InstructReplay has positive full-answer acquisition and higher
\(S_{\mathrm{full}}\) than all seven Table~\ref{tab:results-1} baselines,
BOS replay and Hard in all six settings. Comparator order can change:
under the full criterion the closest Table~\ref{tab:results-1} method is
RAFT on CommonCorpus Qwen, SDFT on CommonCorpus Phi and LoRA-SFT on PMC
Smol. On CommonCorpus Smol, Hard's primary-score lead reverses under full
scoring (\(4.81\) versus \(4.91\) pp).

\textbf{Judging stages.} Table~\ref{tab:results-6} separates vote-slot
outcomes from final question judgments. The same 138 ineligible
CommonCorpus questions occur for every endpoint: pooled over three
models, each method has 414 fixed exclusions and 3,366 issued slots. PMC
has no fixed exclusions and 4,608 issued slots. Each method has 1,536
question outcomes per domain. Invalid slots fail record validation;
valid UNCLEAR-containing slots describe content uncertainty and can
overlap with negative core votes. Core U combines fixed ineligibility
with eligible questions lacking a resolved majority.

\begin{table*}[!tbp]
\centering
\small
\setlength{\tabcolsep}{3pt}
\begin{tabular*}{\textwidth}{@{\extracolsep{\fill}}lrrrrrrr@{}}
\toprule
Method & Invalid slots & Unclear slots & Negative slots & Core yes & Core no & Core U & Core \& Extra \\
\midrule
\multicolumn{8}{l}{\textit{CommonCorpus}} \\
\addlinespace[2pt]
M0 & 361 & 34 & 2567 & 121 & 889 & 526 & 82 \\
Naive SFT & 46 & 83 & 2455 & 254 & 843 & 439 & 66 \\
LoRA-SFT & 45 & 85 & 2782 & 144 & 950 & 442 & 48 \\
Self-Rewrite & 120 & 48 & 2429 & 246 & 835 & 455 & 149 \\
SDFT & 84 & 98 & 2465 & 234 & 841 & 461 & 102 \\
DFT & 38 & 77 & 2848 & 132 & 972 & 432 & 26 \\
RAFT & 133 & 60 & 2319 & 283 & 794 & 459 & 143 \\
Domain-KL & 515 & 54 & 2044 & 246 & 748 & 542 & 65 \\
BOS replay & 518 & 44 & 2029 & 255 & 739 & 542 & 53 \\
Hard & 510 & 44 & 2039 & 248 & 746 & 542 & 54 \\
\fullrowcolor{oursrow}{\textwidth}InstructReplay & 50 & 74 & 2460 & 252 & 840 & 444 & 64 \\
\midrule
\multicolumn{8}{l}{\textit{PMC}} \\
\addlinespace[2pt]
M0 & 578 & 38 & 3314 & 235 & 1175 & 126 & 106 \\
Naive SFT & 78 & 103 & 2697 & 570 & 928 & 38 & 35 \\
LoRA-SFT & 94 & 101 & 3276 & 369 & 1134 & 33 & 13 \\
Self-Rewrite & 162 & 65 & 2884 & 501 & 994 & 41 & 94 \\
SDFT & 119 & 97 & 3228 & 384 & 1117 & 35 & 40 \\
DFT & 102 & 108 & 3439 & 313 & 1179 & 44 & 13 \\
RAFT & 81 & 79 & 2729 & 569 & 937 & 30 & 85 \\
Domain-KL & 91 & 106 & 2762 & 541 & 961 & 34 & 19 \\
BOS replay & 89 & 92 & 2675 & 579 & 924 & 33 & 34 \\
Hard & 86 & 107 & 2695 & 570 & 932 & 34 & 37 \\
\fullrowcolor{oursrow}{\textwidth}InstructReplay & 67 & 93 & 2704 & 580 & 929 & 27 & 37 \\
\bottomrule
\end{tabular*}
\caption{Judging-stage counts pooled over the three models. Slot columns and question columns have different denominators, specified in the text. Core yes/no/U partition the 1,536 questions.  Core \& Extra is their joint majority-confirmation event and can overlap with other columns.  Shaded rows mark InstructReplay.}
\label{tab:results-6}
\end{table*}

Invalid rates differ by method. In CommonCorpus, InstructReplay and BOS
have 50 and 518 invalid slots out of 3,366, respectively; their eligible
unresolved questions number 30 and 128. Confirmation gains can therefore
reflect both answer content and assessability under this rubric.

The targeted CommonCorpus check compares InstructReplay, BOS and
Domain-KL using saved run records. Primary rubric/schema identities and
within-model request settings agree.  We write IR for InstructReplay. In the final labels, BOS has 542 Core U outcomes, comprising 414 fixed exclusions and 128
eligible unresolved questions.

Holding resolved labels, fixed exclusions and F unchanged, the
domain-mean score difference has the deterministic unknown-only range
\([\Delta\bar S-100u_b/1536,\Delta\bar S+100u_{\mathrm{IR}}/1536]\),
where \(u\) counts eligible unresolved questions. For IR minus BOS, this
is \([-1.58,8.71]\) pp on CommonCorpus and \([4.41,8.32]\) pp on PMC. Hard also has 128 eligible unresolved CommonCorpus questions, and 34 on PMC against 27 for IR, so
IR minus Hard has the range \([-7.57,2.72]\) pp on CommonCorpus and
\([-0.69,3.28]\) pp on PMC. These bounds vary only unresolved labels; the supplement's verification script recomputes them.

\textbf{Eligible-denominator sensitivity.} For CommonCorpus,
\(A_{512}=(374/512)A_{374}\), so \(S_{512}=(374/512)A_{374}-F\).
Changing the denominator rescales acquisition relative to forgetting.
Table~\ref{tab:results-7} reports scores over the 374 eligible
questions. InstructReplay still exceeds all seven
Table~\ref{tab:results-1} baselines and BOS in the three settings, while
the closest comparator for Qwen changes to RAFT.

\begin{table*}[!tbp]
\centering
\small
\setlength{\tabcolsep}{3pt}
\begin{tabular*}{\textwidth}{@{\extracolsep{\fill}}lrrrrrr@{}}
\toprule
Model & IR \(S_{512}\) & IR \(S_{374}\) & Best baseline & Baseline \(S_{374}\) & \(\Delta S\) & \(\Delta S\) vs.~BOS \\
\midrule
SmolLM3-3B & 1.01 & \textbf{3.10} & Self-Rewrite & -3.95 & 7.05 & 0.70 \\
Qwen3-4B & 8.56 & \textbf{12.23} & RAFT & 10.77 & 1.46 & 8.83 \\
Phi-4-mini & 7.22 & \textbf{10.89} & Self-Rewrite & 5.30 & 5.59 & 10.51 \\
\bottomrule
\end{tabular*}
\caption{CommonCorpus core-score denominator sensitivity (pp), retaining F. Best baseline excludes M0, InstructReplay and the BOS/Hard controls. Hard's primary-score lead on CommonCorpus SmolLM3-3B persists on this scale. Within each row, bold marks the higher $S_{374}$ of the two shown methods.}
\label{tab:results-7}
\end{table*}

Table~\ref{tab:results-8} contrasts two saved answers to the same evaluation question: one is core-confirmed with unsupported additions,
while the other has an unresolved majority. These examples illustrate
how the rubric separates answer quality from judging uncertainty.

\begin{table*}[!tbp]
\centering
\small
\setlength{\tabcolsep}{3pt}
\begin{tabularx}{\textwidth}{@{}>{\raggedright\arraybackslash}X>{\raggedright\arraybackslash}X>{\raggedright\arraybackslash}X@{}}
\toprule
Observed answer & Core/full votes (Qwen; Gemma; GPT-OSS) & Interpretation \\
\midrule
Smol InstructReplay: ``pmPLGA@SV treatment reduced cerebral infarct volume by 50\% compared to free simvastatin.'' & 1/0; 1/0; U/U & Core confirmed; full not confirmed. Two judges mark the added 50\% as unsupported. \\
Smol Domain-KL: ``pmPLGA@SV treatment significantly reduced cerebral infarct volume in the in vivo model.'' & U/U; U/U; 1/1 & Unconfirmed: Qwen has a schema failure, Gemma marks CORE unclear, GPT-OSS supports it. The majority remains unresolved. \\
\bottomrule
\end{tabularx}
\caption{Two saved answers illustrating core/full confirmation and unresolved judging.  U denotes unknown.}
\label{tab:results-8}
\end{table*}

\subsection{General scoring}\label{app:general}

\paragraph{Evaluation subsets.}
We use fixed subsets to limit the cost of repeated generation and scoring
across models, methods and checkpoints. MMLU contains five test questions from
each of its 57 subjects (285 total); its five-shot demonstrations come from MMLU's dev split. MATH-500 contains 20 problems from
each of five difficulty levels (100 total). BBH contains ten questions from
each of 27 tasks (270 total). HumanEval+ \citep{liu2023evalplus} contains
64 of its 164 problems. Within each stratum, questions are ranked by a
deterministic SHA-256 hash of a fixed namespace and the question identifier, and
the required number with the lowest hashes is retained. These selectors
do not use model outputs or correctness. IFEval uses the first 120
examples in the frozen evaluation dataset order, selected with the
evaluation harness's limit option and reused unchanged across runs;
this is a fixed prefix, without random or stratified sampling. All
compared models and checkpoints use identical question sets and the same
scoring conventions within each backbone. Reported accuracies refer
to these subsets.

IFEval reports prompt-level strict accuracy. MMLU uses five-shot,
conditional option log likelihood with each model's fixed native chat
template and answer-prefill convention; the same convention applies to
M0 and its adapted checkpoints. MATH-500 uses the versioned answer
extractor and symbolic equivalence checker. BBH uses task-aware answer
extraction independent of the gold label, normalized exact comparison,
and zero credit for invalid output. HumanEval+ executes the generated
function against the EvalPlus base and extended tests; a problem counts as
solved only when both pass. A response that contains a full function
definition is executed after removing common leading indentation. The supplement includes the BBH answer extractor. The scores include sensitivity to answer
formatting and extraction. Net benchmark changes can conceal both newly correct and lost correct questions. All compared checkpoints are scored with the same scorers.

\begin{table*}[!tbp]
\centering
\small
\setlength{\tabcolsep}{3pt}
\begin{tabularx}{\textwidth}{@{}>{\raggedright\arraybackslash}X>{\raggedright\arraybackslash}X>{\raggedright\arraybackslash}X>{\raggedright\arraybackslash}X@{}}
\toprule
Task & Answer protocol & New-token cap & Prompt and scorer \\
\midrule
Domain & Greedy generation & 512 & One user turn, closed-book domain question; core/full rubric \\
IFEval & Greedy generation & 1,280 & Frozen instruction text; prompt-level strict scorer \\
MATH-500 & Greedy generation & 1,024 & Frozen problem prompt; versioned extraction and symbolic comparison \\
BBH & Greedy generation & 1,024 & Three-shot CoT prompt; task-aware extraction and exact comparison \\
HumanEval+ & Greedy generation & 768 & Frozen function prompt; EvalPlus base and extended tests \\
MMLU & Five-shot option likelihood & N/A & Native template and model-specific Answer prefill; conditional option scores \\
\bottomrule
\end{tabularx}
\caption{Evaluation decoding, separate from replay generation. Generation uses \texttt{do\_sample=False}, native chat templates, \texttt{enable\_thinking=False}, evaluation mode and BF16. Temperature and top-p do not apply to this greedy path. No system message or model-answer retry is added by the wrapper; task instructions are in the user prompt. Smol/Qwen stop IDs are 128012/151645; Phi uses {[}200020, 199999{]}.  The model inventory provides tokenizer/template hashes and scorer entrypoints.}
\label{tab:results-9}
\end{table*}

\textbf{Saved-output diagnostics.}

Table~\ref{tab:results-10} separates BBH accepted-correct,
accepted-wrong and invalid-extraction outcomes under the task-aware scorer. Relative to M0, InstructReplay has 24 fewer invalid extractions,
35 more accepted-but-wrong answers and 11 fewer correct answers across
the six settings. Naive SFT has 721 invalid extractions versus M0's 96.
Parser acceptance improves for InstructReplay while answer correctness
falls slightly. Figure~\ref{fig:capability} shows the corresponding benchmark-score
changes.

\begin{table}[!htbp]
\centering
\small
\setlength{\tabcolsep}{3pt}
\begin{tabular*}{\columnwidth}{@{\extracolsep{\fill}}lrrr@{}}
\toprule
Method & \shortstack{Accepted\\correct} & \shortstack{Accepted\\wrong} & \shortstack{Invalid\\extraction} \\
\midrule
M0 & 1154 & 370 & 96 \\
Naive SFT & 571 & 328 & 721 \\
LoRA-SFT & \underline{1041} & 390 & \underline{189} \\
Self-Rewrite & 901 & 356 & 363 \\
SDFT & 918 & 436 & 266 \\
DFT & 386 & 326 & 908 \\
RAFT & 938 & 423 & 259 \\
Domain-KL & 770 & 366 & 484 \\
\fullrowcolor{oursrow}{\columnwidth}InstructReplay & \textbf{1143} & 405 & \textbf{72} \\
\bottomrule
\end{tabular*}
\caption{BBH output categories pooled across the six settings, with 1,620 question--endpoint records per method. BOS replay, Hard and Gold replay are not included in this table. The shaded row is InstructReplay. Bold/underline mark the best/second-best adapted count for accepted-correct answers (more) and invalid extractions (fewer); accepted-wrong counts are not ranked, because they also fall when outputs cannot be parsed. M0 is a reference.}
\label{tab:results-10}
\end{table}

The supplementary scoring examples include hash-selected M0/adapted
pairs and six complete short BBH responses selected by parser-result
category. These examples show how the parser scores each category.

\section{Training and Baseline
Implementation}\label{app:training}

\subsection{Shared optimization}\label{c.1.-shared-optimization}

The training schedule is 300 optimizer steps with 32 effective domain
sequences per step. The matched replay runs use eight devices, one
sequence per device per microbatch and four accumulation steps. The
released run-level matrix lists all 48 layouts. CommonCorpus SDFT uses
four devices with accumulation eight; its other methods use eight with
accumulation four. PMC SDFT uses the four-device layout, Self-Rewrite
uses four or eight depending on the model, and the other methods use
eight. Each microbatch contains one sequence per device, giving the same
effective batch of 32. All paths use BF16. The maximum sequence length
is 2,048 for SmolLM3 and Phi, 2,304 for every CommonCorpus Qwen path,
and 2,304 for PMC Qwen except Naive SFT, DFT, and LoRA-SFT, which use 2,048. The
sequence limits accommodate the respective encoded inputs.
InstructReplay uses a constant \(10^{-5}\) learning rate with no warmup,
response-only losses, and 32 additional retention sequences per step.
Student and teacher share the exact initial model, tokenizer, and
chat-template identity; the teacher stays frozen. AdamW uses betas (0.9,
0.999), epsilon \(10^{-8}\), zero weight decay, gradient clipping at 1,
and ZeRO-2. Equal-size rank batches and per-sequence losses determine
the accumulation reduction. Samples are right-padded without packing;
prompt/padding labels are masked. Attention dropout is zero in the model
configurations, and LoRA dropout is zero. Student forwards use training
mode; frozen-reference forwards use evaluation mode without gradients.
Non-reentrant gradient checkpointing is enabled. Tokenizer and template
identities are bound with the model tree.

\subsection{Baseline targets and
adaptation}\label{app:baselines}

\begin{table*}[!tbp]
\centering\small
\setlength{\tabcolsep}{4pt}
\begin{tabularx}{\textwidth}{@{}>{\raggedright\arraybackslash}p{0.18\textwidth}>{\raggedright\arraybackslash}X>{\raggedright\arraybackslash}X>{\raggedright\arraybackslash}X@{}}
\toprule
Method & Training input and target & Reference / teacher & Training objective / main change \\
\midrule
Naive SFT & Domain question and supplied answer & None & Full-parameter CE \\
\addlinespace[3pt]
LoRA-SFT & Same domain pairs & None & Rank-16 adapters; learning rate \(10^{-4}\) \\
\addlinespace[3pt]
Self-Rewrite & Domain question and M0 rewrite; original-answer fallback & Initial model for offline rewriting & CE on fixed rewritten answers \\
\addlinespace[3pt]
SDFT & Current-student domain responses with demonstrations & Teacher EMA, updated each optimizer step & Full-vocabulary forward KL, temperature 1 \\
\addlinespace[3pt]
DFT & Domain question and supplied answer & Detached current-student probability & Target-token probability weights CE \\
\addlinespace[3pt]
RAFT & Refined answer and student domain responses & Frozen M0 with answer-conditioned context & CE plus top-k forward KL; rank-local loss EMA ratio \\
\addlinespace[3pt]
Domain-KL & Domain gold-answer prefixes & Frozen initial model & CE plus 0.25 forward KL on domain prefixes \\
\addlinespace[3pt]
InstructReplay & Domain pairs plus saved general-instruction responses & Frozen initial model & Domain CE plus 0.25 forward KL on instruction prefixes \\
\addlinespace[3pt]
BOS replay & Domain pairs plus BOS-generated responses & Frozen initial model & Same replay objective; BOS contexts and matched response tokens \\
\addlinespace[3pt]
Hard & Domain pairs plus the same instruction responses as InstructReplay & No teacher in the optimized replay loss & Domain CE plus 0.25 next-token CE on replay \\
\addlinespace[3pt]
Gold replay & Domain pairs plus the collection's own answers to the same instructions & None & Domain CE plus 0.25 next-token CE on replay \\
\addlinespace[3pt]
\bottomrule\end{tabularx}
\caption{Baseline configuration matrix: inputs, targets and references under the shared training protocol. Method-specific adaptations are described in Appendix~\ref{app:baselines}. BOS replay, Hard and Gold replay are controlled variants of InstructReplay that change only the replay context, the target or the source of the replayed answers; they are not tuned baselines.}\label{tab:baseline-configuration}\label{tab:results-11}
\end{table*}

LoRA-SFT uses rank 16, alpha 32, zero dropout, and no bias adaptation.
It targets attention q/k/v/o projections and feed-forward gate/up/down
projections, with learning rate \(10^{-4}\). Naive SFT uses the original
domain answers and full parameters. DFT uses detached target-token
probabilities to weight negative log likelihood, using within-sequence
means.

Self-Rewrite uses the exact M0 to rewrite domain answers with access to
the gold answer, applies a generic-answer filter with fallback to the
original, and trains full parameters on the resulting supervision. SDFT
follows the practical forward-KL choice described in v2 of
\citet{shenfeld2026sdft}. It uses current-student responses and a
demonstration-conditioned teacher updated once per optimizer step as
\(\theta_T\leftarrow0.99\theta_T+0.01\theta\). Its loss is
full-vocabulary teacher-to-student KL at temperature 1; generation is
capped at 512 tokens within the total sequence limit. It has no separate
general replay population.

RAFT follows \citet{li2026raft}. It obtains an M0 rewrite, applies a
cosine-similarity threshold of 0.8 using all-MiniLM-L6-v2, and uses
Qwen3.8-27B for answer fusion on the retained path. Three-judge quality
scoring stops fusion early at 0.8; if no candidate reaches that
threshold, the highest-scoring candidate is still selected. Online
training combines supervision with answer-conditioned reference feedback
on up to 64 student-generated tokens. KL uses the teacher's top 512
tokens at temperature 1.5, normalized over that support.

Adaptive balancing uses the ratio of detached CE and KL loss EMAs:
\(w=(m_{\mathrm{CE}}/b)/(m_{\mathrm{KL}}/b+10^{-8})\), where
\(b=1-0.9^{s+1}\). The rank-local 0.9 EMA is updated per microbatch. Its
common correction factor \(1-0.9^{s+1}\) uses optimizer step \(s\) and
cancels in the CE/KL ratio except for the epsilon term. The rank-local
update makes balancing dependent on the recorded rank/microbatch layout.

Domain-KL uses frozen-reference forward KL on domain gold-answer
prefixes, with full vocabulary, temperature 1, and fixed coefficient
0.25. Following the reference-matching principle of
\citet{yang2024annealing}, this adaptation keeps the coefficient fixed
throughout training.  Across the six
settings, Self-Rewrite retains at least 97.5\% of rewrites and falls back
to the original answer otherwise; its final mean response length varies from 40 to 142 tokens across settings. RAFT takes the fusion branch for 60\% to 73\% of examples, with final mean response lengths of 26 to 57 tokens. Among
accepted Self-Rewrite outputs, a lexical numeric-anchor scan flags
248/312/342 CommonCorpus rows and 26/68/572 PMC rows (Smol/Qwen/Phi)
with a missing numeric anchor. The following sample inspects
numeric-string preservation in more detail.

A fixed rule selects 32 numeric-bearing training questions per domain,
including eight with numeric-unit expressions, using only gold answers
and hashed task IDs. The same questions are checked across three models
and both Self-Rewrite and RAFT. Across the six settings, Self-Rewrite
retains all source numeric literals in 178/192 checked targets and all
selected number--unit expressions in 43/48; RAFT retains them in 189/192
and 47/48. This sample checks literal preservation with
limited normalization. Unit equivalence and the surrounding factual
relations require semantic assessment.

The public source examples also retain all 12 method/model targets on
the two illustrated facts. They expose distinctions missed by lexical
retention: CommonCorpus Smol Self-Rewrite expands INGP as ``Neural Image
Generation Pipeline'', which the fixed evidence does not supply; PMC
Smol RAFT adds ``30\% to 50\%'', while the fixed evidence gives no such
range.

\subsection{Validation coverage}\label{c.3.-validation-coverage}

Forward KL uses FP32 log probabilities and its analytical logit gradient
\(q-q_0\), avoiding a finite-precision residual at identical logits. A
masked, unequal-length CPU comparison with an FP64 autograd oracle gives
maximum gradient error \(1.84\times10^{-8}\); an eight-rank nonzero-KL
check gives gradient cosine 0.999954 and relative L2 error 0.009600.

All 48 trained endpoints completed 300 steps, and each restored checkpoint passes a reload check: 39 by a paired save/reload comparison and nine by reproducing fixed-input logits and short greedy outputs across two independent restores.

All three instruction-replay buffers retain 9,600 prompts without
rejection or replacement. The maximum Qwen replay length is 2,082
tokens, within its actual 2,304-token limit. The run-level length matrix
confirms that all 42 static-supervision sets fit their recorded limits, and the six on-policy SDFT runs generate their targets during training.

\section{Complete Scores}\label{app:scores}

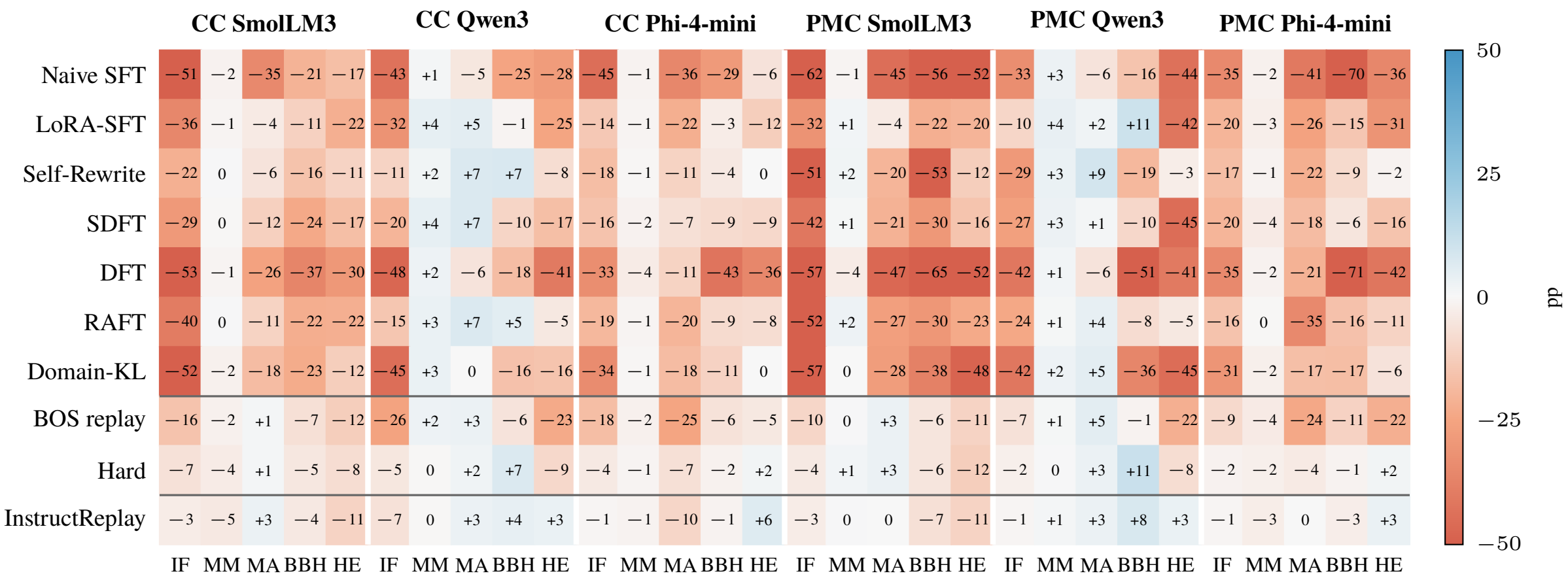
\begin{figure*}[!tbp]
\centering
\begin{tikzpicture}
\begin{axis}[scale only axis, width=0.79\textwidth, height=5.0cm, enlargelimits=false, clip=false,
  xmin=-0.5, xmax=29.5, ymin=-0.5, ymax=9.5, y dir=reverse, axis line style={draw=none}, tick style={draw=none},
  xtick={0,...,29}, xticklabels={IF,MM,MA,BBH,HE,IF,MM,MA,BBH,HE,IF,MM,MA,BBH,HE,IF,MM,MA,BBH,HE,IF,MM,MA,BBH,HE,IF,MM,MA,BBH,HE},
  xticklabel style={font=\tiny}, ytick={0,...,9},
  yticklabels={Naive SFT,LoRA-SFT,Self-Rewrite,SDFT,DFT,RAFT,Domain-KL,BOS replay,Hard,InstructReplay},
  yticklabel style={font=\scriptsize},
  colormap={div}{rgb255=(214,96,77) rgb255=(244,165,130) rgb255=(247,247,247) rgb255=(146,197,222) rgb255=(67,147,195)},
  point meta min=-50, point meta max=50,
  colorbar, colorbar style={width=0.18cm, ytick={-50,-25,0,25,50}, yticklabel style={font=\tiny}, ylabel={pp}, ylabel style={font=\tiny}},
  nodes near coords={\lab}, visualization depends on={value \thisrow{label}\as\lab},
  every node near coord/.append style={font=\fontsize{4.6}{5}\selectfont, anchor=center, inner sep=0pt}]
\addplot[matrix plot*, mesh/cols=30, point meta=explicit] table[x=x, y=y, meta=v] {figures/data/capability.dat};
\draw[white, line width=1.6pt] (axis cs:4.5,-0.5) -- (axis cs:4.5,9.5);
\draw[white, line width=1.6pt] (axis cs:9.5,-0.5) -- (axis cs:9.5,9.5);
\draw[white, line width=1.6pt] (axis cs:14.5,-0.5) -- (axis cs:14.5,9.5);
\draw[white, line width=1.6pt] (axis cs:19.5,-0.5) -- (axis cs:19.5,9.5);
\draw[white, line width=1.6pt] (axis cs:24.5,-0.5) -- (axis cs:24.5,9.5);
\draw[black!60, line width=0.6pt] (axis cs:-0.5,6.5) -- (axis cs:29.5,6.5);
\draw[black!60, line width=0.6pt] (axis cs:-0.5,8.5) -- (axis cs:29.5,8.5);
\node[font=\scriptsize\bfseries, anchor=south] at (axis cs:2,-0.6) {CC SmolLM3};
\node[font=\scriptsize\bfseries, anchor=south] at (axis cs:7,-0.6) {CC Qwen3};
\node[font=\scriptsize\bfseries, anchor=south] at (axis cs:12,-0.6) {CC Phi-4-mini};
\node[font=\scriptsize\bfseries, anchor=south] at (axis cs:17,-0.6) {PMC SmolLM3};
\node[font=\scriptsize\bfseries, anchor=south] at (axis cs:22,-0.6) {PMC Qwen3};
\node[font=\scriptsize\bfseries, anchor=south] at (axis cs:27,-0.6) {PMC Phi-4-mini};
\end{axis}
\end{tikzpicture}
\caption{Signed General-score change relative to $M_0$ for every method and setting (pp, rounded; colors saturate at $\pm50$). IF, MM, MA, BBH and HE denote IFEval, MMLU, MATH-500, BBH and HumanEval+. Rows below the first rule are the replay controls; the last row is InstructReplay. Positive changes are shown even though $F$ counts only declines. CC abbreviates CommonCorpus.}
\label{fig:capability}
\end{figure*}

Figure~\ref{fig:capability} separates the five General benchmarks whose declines enter F.
Tables \ref{tab:cc-full}–\ref{tab:pmc-full} report the raw endpoint
scores.

Tables \ref{tab:cc-full}–\ref{tab:pmc-full} retain all 54 rows. Domain
reports the core-confirmation rate \(100C/N_D\) on a scale of 0 to 100, with
\(N_D=512\) and unknown questions retained in the denominator. IF, MM, MA and
HE denote IFEval, MMLU, MATH-500 and HumanEval+. Domain and General scores are
percentages; A, F, W, and S are percentage points. Domain-KL denotes the
fixed-coefficient domain-prefix adaptation in Appendix~\ref{app:baselines}. Each M0 is
the corresponding model's unchanged checkpoint.

\begin{table*}[!tbp]
\centering
\footnotesize
\setlength{\tabcolsep}{3pt}
\begin{tabularx}{\textwidth}{@{}l*{10}{>{\raggedleft\arraybackslash}X}@{}}
\toprule
 & \multicolumn{1}{c}{Domain (\%)} & \multicolumn{5}{c}{General (\%)} & \multicolumn{4}{c}{Acquisition / forgetting (pp)} \\
\cmidrule(lr){2-2}\cmidrule(lr){3-7}\cmidrule(lr){8-11}
Method & Core $\uparrow$ & IF & MM & MA & BBH & HE & $A\uparrow$ & $F\downarrow$ & $W\downarrow$ & $S\uparrow$ \\
\midrule
\multicolumn{11}{l}{\textit{SmolLM3-3B}} \\
\addlinespace[2pt]
M0 & 9.77 & 70.83 & 59.30 & 71.00 & 73.70 & 51.56 & 0.00 & 0.00 & 0.00 & 0.00 \\
Naive SFT & \underline{17.38} & 20.00 & 57.54 & 36.00 & 52.96 & 34.38 & \underline{7.62} & 25.10 & 50.83 & -17.49 \\
LoRA-SFT & 8.59 & 35.00 & 58.25 & \underline{67.00} & \underline{62.22} & 29.69 & -1.17 & 14.85 & 35.83 & -16.02 \\
Self-Rewrite & 14.84 & \underline{49.17} & \textbf{59.65} & 65.00 & 57.78 & \textbf{40.62} & 5.08 & \underline{10.91} & \underline{21.67} & \underline{-5.83} \\
SDFT & 15.43 & 41.67 & \underline{58.95} & 59.00 & 50.00 & 34.38 & 5.66 & 16.48 & 29.17 & -10.82 \\
DFT & 8.40 & 18.33 & 58.25 & 45.00 & 37.04 & 21.88 & -1.37 & 29.18 & 52.50 & -30.55 \\
RAFT & \textbf{18.55} & 30.83 & \textbf{59.65} & 60.00 & 51.48 & 29.69 & \textbf{8.79} & 19.02 & 40.00 & -10.23 \\
Domain-KL & \underline{17.38} & 19.17 & 57.19 & 53.00 & 50.74 & \underline{39.06} & \underline{7.62} & 21.45 & 51.67 & -13.83 \\
\fullrowcolor{oursrow}{\textwidth}InstructReplay & 15.43 & \textbf{67.50} & 54.74 & \textbf{74.00} & \textbf{69.26} & \textbf{40.62} & 5.66 & \textbf{4.66} & \textbf{10.94} & \textbf{1.01} \\
\midrule
\multicolumn{11}{l}{\textit{Qwen3-4B}} \\
\addlinespace[2pt]
M0 & 7.42 & 80.83 & 71.58 & 67.00 & 66.30 & 59.38 & 0.00 & 0.00 & 0.00 & 0.00 \\
Naive SFT & 16.21 & 37.50 & 72.63 & 62.00 & 41.11 & 31.25 & 8.79 & 20.33 & 43.33 & -11.54 \\
LoRA-SFT & 10.35 & 49.17 & \textbf{75.79} & \underline{72.00} & 65.19 & 34.38 & 2.93 & 11.56 & 31.67 & -8.63 \\
Self-Rewrite & \underline{17.97} & \underline{70.00} & 73.33 & \textbf{74.00} & \textbf{73.70} & 51.56 & \underline{10.55} & \underline{3.73} & \underline{10.83} & \underline{6.82} \\
SDFT & 16.21 & 60.83 & \textbf{75.79} & \textbf{74.00} & 55.93 & 42.19 & 8.79 & 9.51 & 20.00 & -0.72 \\
DFT & 9.77 & 33.33 & 74.04 & 61.00 & 48.52 & 18.75 & 2.34 & 22.38 & 47.50 & -20.04 \\
RAFT & \textbf{18.16} & 65.83 & \underline{74.74} & \textbf{74.00} & \underline{71.48} & \underline{54.69} & \textbf{10.74} & 3.94 & 15.00 & 6.80 \\
Domain-KL & 16.21 & 35.83 & \underline{74.74} & 67.00 & 50.00 & 43.75 & 8.79 & 15.38 & 45.00 & -6.60 \\
\fullrowcolor{oursrow}{\textwidth}InstructReplay & 17.38 & \textbf{74.17} & 71.23 & 70.00 & 70.37 & \textbf{62.50} & 9.96 & \textbf{1.40} & \textbf{6.67} & \textbf{8.56} \\
\midrule
\multicolumn{11}{l}{\textit{Phi-4-mini-instruct}} \\
\addlinespace[2pt]
M0 & 6.45 & 68.33 & 71.58 & 69.00 & 73.70 & 56.25 & 0.00 & 0.00 & 0.00 & 0.00 \\
Naive SFT & 16.02 & 23.33 & 70.18 & 33.00 & 44.81 & 50.00 & 9.57 & 23.51 & 45.00 & -13.94 \\
LoRA-SFT & 9.18 & \underline{54.17} & \underline{70.53} & 47.00 & \underline{70.37} & 43.75 & 2.73 & 10.61 & 22.00 & -7.88 \\
Self-Rewrite & 15.23 & 50.83 & \textbf{70.88} & 58.00 & 69.26 & \underline{56.25} & 8.79 & \underline{6.73} & 17.50 & \underline{2.06} \\
SDFT & 14.06 & 52.50 & 69.82 & \textbf{62.00} & 65.19 & 46.88 & 7.62 & 8.50 & \underline{15.83} & -0.88 \\
DFT & 7.62 & 35.00 & 68.07 & 58.00 & 31.11 & 20.31 & 1.17 & 25.27 & 42.59 & -24.10 \\
RAFT & \textbf{18.55} & 49.17 & \underline{70.53} & 49.00 & 65.19 & 48.44 & \textbf{12.11} & 11.31 & 20.00 & 0.80 \\
Domain-KL & 14.45 & 34.17 & \textbf{70.88} & 51.00 & 62.59 & \underline{56.25} & 8.01 & 12.80 & 34.17 & -4.79 \\
\fullrowcolor{oursrow}{\textwidth}InstructReplay & \underline{16.41} & \textbf{67.50} & 70.18 & \underline{59.00} & \textbf{72.22} & \textbf{62.50} & \underline{9.96} & \textbf{2.74} & \textbf{10.00} & \textbf{7.22} \\
\bottomrule
\end{tabularx}
\caption{Complete CommonCorpus results. The denominator includes 374 eligible and 138 fixed ineligible questions (Appendix~\ref{app:scoring}). Within each model block, bold/underline mark the best/second-best adapted result in each column: higher Core, General, $A$ and $S$, lower $F$ and $W$. M0 is a reference; ties at the shown precision share marks. Shaded rows are InstructReplay. Table~\ref{tab:results-18} reports the matched controls.}
\label{tab:results-12}
\label{tab:cc-full}
\end{table*}

\begin{table*}[!tbp]
\centering
\footnotesize
\setlength{\tabcolsep}{3pt}
\begin{tabularx}{\textwidth}{@{}l*{10}{>{\raggedleft\arraybackslash}X}@{}}
\toprule
 & \multicolumn{1}{c}{Domain (\%)} & \multicolumn{5}{c}{General (\%)} & \multicolumn{4}{c}{Acquisition / forgetting (pp)} \\
\cmidrule(lr){2-2}\cmidrule(lr){3-7}\cmidrule(lr){8-11}
Method & Core $\uparrow$ & IF & MM & MA & BBH & HE & $A\uparrow$ & $F\downarrow$ & $W\downarrow$ & $S\uparrow$ \\
\midrule
\multicolumn{11}{l}{\textit{SmolLM3-3B}} \\
\addlinespace[2pt]
M0 & 16.21 & 70.83 & 59.30 & 71.00 & 73.70 & 51.56 & 0.00 & 0.00 & 0.00 & 0.00 \\
Naive SFT & \underline{37.11} & 9.17 & 58.25 & 26.00 & 18.15 & 0.00 & \underline{20.90} & 42.97 & 61.67 & -22.07 \\
LoRA-SFT & 21.88 & \underline{39.17} & 60.70 & \underline{67.00} & \underline{51.85} & 31.25 & 5.66 & \underline{15.57} & \underline{31.67} & -9.90 \\
Self-Rewrite & 31.64 & 20.00 & \underline{61.05} & 51.00 & 21.11 & \underline{39.06} & 15.43 & 27.19 & 52.59 & -11.76 \\
SDFT & 27.15 & 29.17 & 60.00 & 50.00 & 44.07 & 35.94 & 10.94 & 21.58 & 41.67 & -10.65 \\
DFT & 18.75 & 14.17 & 55.79 & 24.00 & 8.52 & 0.00 & 2.54 & 44.78 & 65.19 & -42.25 \\
RAFT & 36.72 & 19.17 & \textbf{61.75} & 44.00 & 43.70 & 28.12 & 20.51 & 26.42 & 51.67 & \underline{-5.91} \\
Domain-KL & 35.16 & 14.17 & 59.30 & 43.00 & 35.56 & 3.12 & 18.95 & 34.25 & 56.67 & -15.31 \\
\fullrowcolor{oursrow}{\textwidth}InstructReplay & \textbf{38.67} & \textbf{67.50} & 59.65 & \textbf{71.00} & \textbf{66.67} & \textbf{40.62} & \textbf{22.46} & \textbf{4.26} & \textbf{10.94} & \textbf{18.20} \\
\midrule
\multicolumn{11}{l}{\textit{Qwen3-4B}} \\
\addlinespace[2pt]
M0 & 18.55 & 80.83 & 71.58 & 67.00 & 66.30 & 59.38 & 0.00 & 0.00 & 0.00 & 0.00 \\
Naive SFT & \underline{37.11} & 47.50 & 74.39 & 61.00 & 50.74 & 15.62 & \underline{18.55} & 19.73 & 43.75 & -1.17 \\
LoRA-SFT & 28.12 & \underline{70.83} & \textbf{75.44} & 69.00 & \textbf{77.04} & 17.19 & 9.57 & 10.44 & 42.19 & -0.87 \\
Self-Rewrite & \textbf{37.30} & 51.67 & \underline{74.74} & \textbf{76.00} & 47.04 & \underline{56.25} & \textbf{18.75} & 10.31 & 29.17 & 8.44 \\
SDFT & 25.59 & 54.17 & 74.39 & 68.00 & 56.67 & 14.06 & 7.03 & 16.32 & 45.31 & -9.29 \\
DFT & 20.12 & 39.17 & 72.98 & 61.00 & 15.56 & 18.75 & 1.56 & 27.81 & 50.74 & -26.24 \\
RAFT & 35.74 & 56.67 & 72.28 & 71.00 & 58.15 & 54.69 & 17.19 & \underline{7.40} & \underline{24.17} & \underline{9.79} \\
Domain-KL & 35.55 & 39.17 & 74.04 & \underline{72.00} & 30.00 & 14.06 & 16.99 & 24.66 & 45.31 & -7.66 \\
\fullrowcolor{oursrow}{\textwidth}InstructReplay & \textbf{37.30} & \textbf{80.00} & 72.63 & 70.00 & \underline{74.07} & \textbf{62.50} & \textbf{18.75} & \textbf{0.17} & \textbf{0.83} & \textbf{18.58} \\
\midrule
\multicolumn{11}{l}{\textit{Phi-4-mini-instruct}} \\
\addlinespace[2pt]
M0 & 11.13 & 68.33 & 71.58 & 69.00 & 73.70 & 56.25 & 0.00 & 0.00 & 0.00 & 0.00 \\
Naive SFT & 37.11 & 33.33 & 69.47 & 28.00 & 3.70 & 20.31 & 25.98 & 36.81 & 70.00 & -10.83 \\
LoRA-SFT & 22.07 & 48.33 & 68.77 & 43.00 & 58.89 & 25.00 & 10.94 & 18.97 & 31.25 & -8.04 \\
Self-Rewrite & 28.91 & 50.83 & \underline{70.18} & 47.00 & 64.81 & \underline{54.69} & 17.77 & \underline{10.27} & 22.00 & 7.50 \\
SDFT & 22.27 & 48.33 & 67.72 & 51.00 & \underline{68.15} & 40.62 & 11.13 & 12.61 & \underline{20.00} & -1.48 \\
DFT & 22.27 & 33.33 & 69.82 & 48.00 & 2.22 & 14.06 & 11.13 & 34.28 & 71.48 & -23.15 \\
RAFT & \textbf{38.67} & \underline{52.50} & \textbf{71.23} & 34.00 & 57.41 & 45.31 & \textbf{27.54} & 15.68 & 35.00 & \underline{11.86} \\
Domain-KL & 34.96 & 37.50 & 69.12 & \underline{52.00} & 56.30 & 50.00 & 23.83 & 14.79 & 30.83 & 9.04 \\
\fullrowcolor{oursrow}{\textwidth}InstructReplay & \underline{37.30} & \textbf{67.50} & 68.42 & \textbf{69.00} & \textbf{70.74} & \textbf{59.38} & \underline{26.17} & \textbf{1.39} & \textbf{3.16} & \textbf{24.78} \\
\bottomrule
\end{tabularx}
\caption{Complete PMC biomedical results, using the metric definitions of Table~\ref{tab:cc-full}. Within each model block, bold/underline mark the best/second-best adapted result in each column, with directions and ties as in Table~\ref{tab:cc-full}; M0 is a reference. Shaded rows are InstructReplay. Cross-domain differences also reflect source genre, chronology, fact structure, and initial model performance.}
\label{tab:results-13}
\label{tab:pmc-full}
\end{table*}

\section{Weight Ablation and Training
Cost}\label{app:lambda}

\subsection{Retention-loss
weight}\label{e.1.-retention-loss-weight}

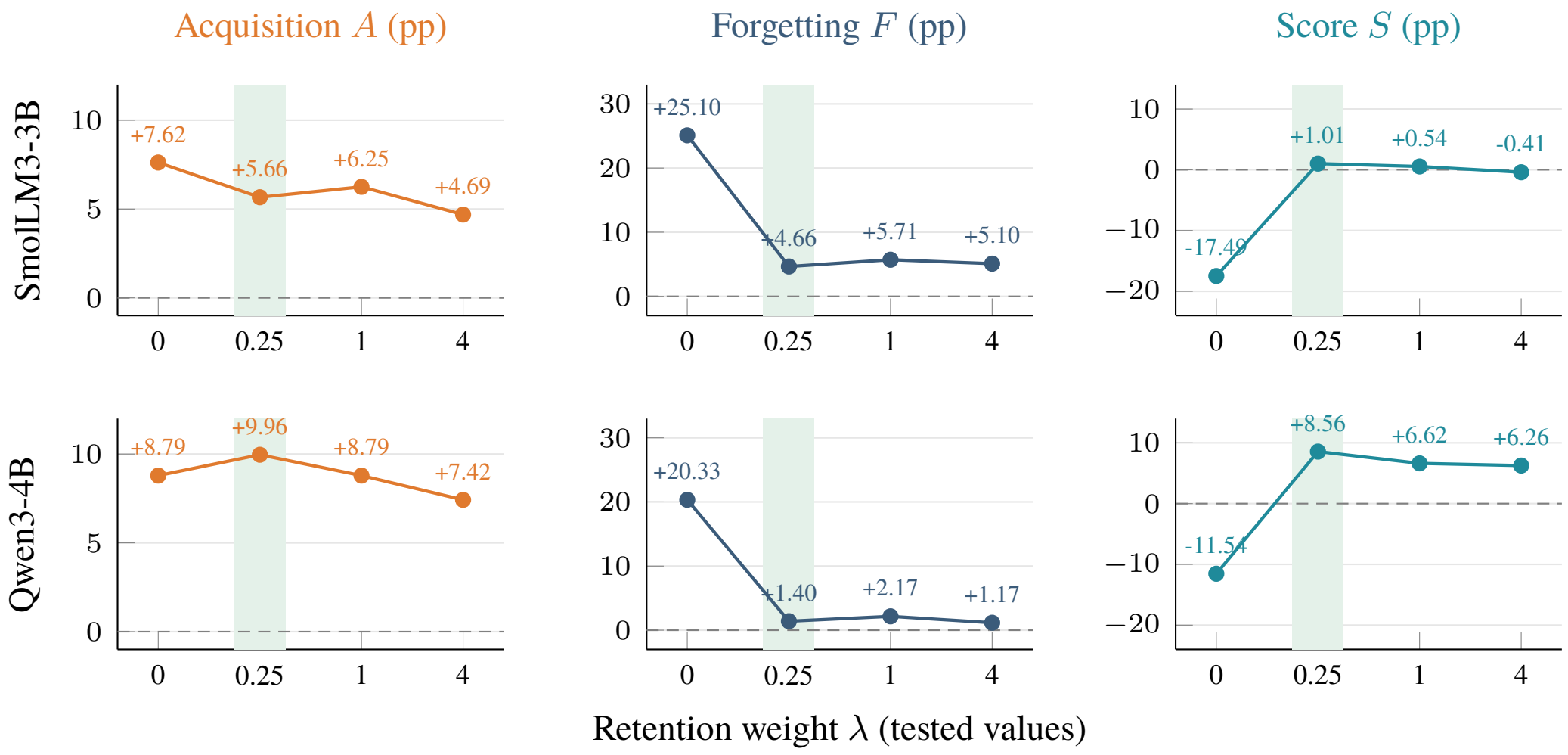
\begin{figure*}[!tbp]
\centering
\begin{tikzpicture}
\begin{groupplot}[group style={group size=3 by 2, horizontal sep=1.25cm, vertical sep=0.9cm},
  width=0.31\textwidth, height=3.6cm, xmin=-0.4, xmax=3.4, xtick={0,1,2,3}, xticklabels={0,0.25,1,4},
  ticklabel style={font=\scriptsize}, title style={font=\small}, axis lines*=left, ymajorgrids, grid style={gray!20},
  every node near coord/.append style={font=\tiny, anchor=south, yshift=1pt}, nodes near coords,
  point meta=explicit symbolic]
\nextgroupplot[title={Acquisition $A$ (pp)}, title style={color=acqcol}, ylabel={SmolLM3-3B}, ylabel style={font=\small}, ymin=-1, ymax=12]
\fill[ircol!12] (axis cs:0.75,-1) rectangle (axis cs:1.25,12);
\draw[densely dashed, gray] (axis cs:-0.4,0) -- (axis cs:3.4,0);
\addplot[acqcol, thick, mark=*, mark size=1.6pt] table[x=x, y=A, meta=Alab] {figures/data/lambda-smollm3.dat};
\nextgroupplot[title={Forgetting $F$ (pp)}, title style={color=forcol}, ymin=-3, ymax=33]
\fill[ircol!12] (axis cs:0.75,-3) rectangle (axis cs:1.25,33);
\draw[densely dashed, gray] (axis cs:-0.4,0) -- (axis cs:3.4,0);
\addplot[forcol, thick, mark=*, mark size=1.6pt] table[x=x, y=F, meta=Flab] {figures/data/lambda-smollm3.dat};
\nextgroupplot[title={Score $S$ (pp)}, title style={color=retcol}, ymin=-24, ymax=14]
\fill[ircol!12] (axis cs:0.75,-24) rectangle (axis cs:1.25,14);
\draw[densely dashed, gray] (axis cs:-0.4,0) -- (axis cs:3.4,0);
\addplot[retcol, thick, mark=*, mark size=1.6pt] table[x=x, y=S, meta=Slab] {figures/data/lambda-smollm3.dat};
\nextgroupplot[ylabel={Qwen3-4B}, ylabel style={font=\small}, ymin=-1, ymax=12]
\fill[ircol!12] (axis cs:0.75,-1) rectangle (axis cs:1.25,12);
\draw[densely dashed, gray] (axis cs:-0.4,0) -- (axis cs:3.4,0);
\addplot[acqcol, thick, mark=*, mark size=1.6pt] table[x=x, y=A, meta=Alab] {figures/data/lambda-qwen3_4b.dat};
\nextgroupplot[ymin=-3, ymax=33, xlabel={Retention weight $\lambda$ (tested values)}, xlabel style={font=\small}]
\fill[ircol!12] (axis cs:0.75,-3) rectangle (axis cs:1.25,33);
\draw[densely dashed, gray] (axis cs:-0.4,0) -- (axis cs:3.4,0);
\addplot[forcol, thick, mark=*, mark size=1.6pt] table[x=x, y=F, meta=Flab] {figures/data/lambda-qwen3_4b.dat};
\nextgroupplot[ymin=-24, ymax=14]
\fill[ircol!12] (axis cs:0.75,-24) rectangle (axis cs:1.25,14);
\draw[densely dashed, gray] (axis cs:-0.4,0) -- (axis cs:3.4,0);
\addplot[retcol, thick, mark=*, mark size=1.6pt] table[x=x, y=S, meta=Slab] {figures/data/lambda-qwen3_4b.dat};
\end{groupplot}
\end{tikzpicture}
\caption{CommonCorpus sensitivity to retention weight at a fixed 1:1 sequence ratio and 300-step schedule. Lines connect the four tested weights on a categorical axis. Each column isolates one metric and shares its vertical scale across models. Every value is labeled; shading marks the default \(\lambda=0.25\). The dashed zero lines mark \(M_0\)'s metric values (\(A=F=S=0\)).}
\label{fig:lambda}
\end{figure*}

\begin{table*}[!tbp]
\centering
\footnotesize
\setlength{\tabcolsep}{3pt}
\begin{tabularx}{\textwidth}{@{}l*{10}{>{\raggedleft\arraybackslash}X}@{}}
\toprule
 & \multicolumn{1}{c}{Domain (\%)} & \multicolumn{5}{c}{General (\%)} & \multicolumn{4}{c}{Acquisition / forgetting (pp)} \\
\cmidrule(lr){2-2}\cmidrule(lr){3-7}\cmidrule(lr){8-11}
Weight $\lambda$ & Core $\uparrow$ & IF & MM & MA & BBH & HE & $A\uparrow$ & $F\downarrow$ & $W\downarrow$ & $S\uparrow$ \\
\midrule
\multicolumn{11}{l}{\textit{SmolLM3-3B}} \\
\addlinespace[2pt]
M0 & 9.77 & 70.83 & 59.30 & 71.00 & 73.70 & 51.56 & 0.00 & 0.00 & 0.00 & 0.00 \\
0 & \textbf{17.38} & 20.00 & \textbf{57.54} & 36.00 & 52.96 & 34.38 & \textbf{7.62} & 25.10 & 50.83 & -17.49 \\
\rowcolor{oursrow}
0.25 & 15.43 & \textbf{67.50} & 54.74 & \textbf{74.00} & \textbf{69.26} & \underline{40.62} & 5.66 & \textbf{4.66} & \underline{10.94} & \textbf{1.01} \\
1 & \underline{16.02} & 63.33 & 54.74 & \underline{72.00} & \underline{68.15} & \underline{40.62} & \underline{6.25} & 5.71 & \underline{10.94} & \underline{0.54} \\
4 & 14.45 & \underline{64.17} & \underline{55.79} & \textbf{74.00} & 67.78 & \textbf{42.19} & 4.69 & \underline{5.10} & \textbf{9.38} & -0.41 \\
\midrule
\multicolumn{11}{l}{\textit{Qwen3-4B}} \\
\addlinespace[2pt]
M0 & 7.42 & 80.83 & 71.58 & 67.00 & 66.30 & 59.38 & 0.00 & 0.00 & 0.00 & 0.00 \\
0 & \underline{16.21} & 37.50 & \textbf{72.63} & 62.00 & 41.11 & 31.25 & \underline{8.79} & 20.33 & 43.33 & -11.54 \\
\rowcolor{oursrow}
0.25 & \textbf{17.38} & \underline{74.17} & 71.23 & \textbf{70.00} & \underline{70.37} & 62.50 & \textbf{9.96} & \underline{1.40} & \underline{6.67} & \textbf{8.56} \\
1 & \underline{16.21} & 70.00 & \textbf{72.63} & \underline{68.00} & \textbf{70.74} & \underline{64.06} & \underline{8.79} & 2.17 & 10.83 & \underline{6.62} \\
4 & 14.84 & \textbf{75.00} & \underline{72.28} & \textbf{70.00} & 67.04 & \textbf{67.19} & 7.42 & \textbf{1.17} & \textbf{5.83} & 6.26 \\
\bottomrule
\end{tabularx}
\caption{CommonCorpus loss-weight ablation, grouped by backbone, with fixed sequence ratio and schedule. The shaded row is the default weight \(\lambda=0.25\). M0, weight 0, and weight 0.25 agree with the corresponding reference, Naive SFT, and InstructReplay rows. All endpoints use the current scoring protocol. Within each backbone, bold/underline mark the best/second-best tested weight in every metric, with directions and ties as in Table~\ref{tab:cc-full}; M0 is a reference.}
\label{tab:results-14}
\end{table*}

The ablation reuses saved checkpoints and model answers. The four
additional configurations for two CommonCorpus backbones at
\(\lambda\in\{1,4\}\) are evaluated with the same domain and general scorers.

\subsection{Training cost}\label{e.2.-training-cost}

\begin{table}[!htbp]
\centering
\small
\setlength{\tabcolsep}{3pt}
\begin{tabularx}{\columnwidth}{@{}l*{3}{>{\raggedleft\arraybackslash}X}@{}}
\toprule
Method & \shortstack{SmolLM3\\GPUh} & \shortstack{Qwen3\\GPUh} & \shortstack{Phi-4-mini\\GPUh} \\
\midrule
LoRA-SFT & 1.19 & 1.40 & \underline{0.87}$^{\dagger}$ \\
Self-Rewrite & \textbf{0.79} & \textbf{0.96} & \textbf{0.67} \\
SDFT & 9.07 & 5.17 & 3.58 \\
DFT & \underline{0.82} & \underline{1.08} & 1.04$^{\dagger}$ \\
RAFT & 6.30 & 7.87 & 4.83 \\
Domain-KL & 1.44 & 1.90 & 1.32 \\
\rowcolor{oursrow}
InstructReplay & 1.64 & 2.22 & 1.91 \\
\bottomrule
\end{tabularx}
\caption{Allocated GPU-hours during successful step-300 PMC training. Device types are listed separately. Offline preparation, failed attempts and evaluation add to these costs. A100 unless marked $^{\dagger}$ (A800). The shaded row is InstructReplay. Bold/underline mark the lowest/second-lowest recorded GPU-hours per backbone, with hardware as noted.}
\label{tab:results-15}
\end{table}

InstructReplay's listed training costs are 1.64, 2.22, and 1.91
GPU-hours in model order SmolLM3, Qwen3, and Phi. RAFT's
corresponding costs are 6.30, 7.87, and 4.83. These measurements cover
the training loop. Retention generation, answer rewriting/fusion, failed
runs and evaluation are separate costs. Naive SFT training time was not
included in the extracted PMC cost records and is not imputed.

Training exposure is 9,600 domain sequences per method. In Smol/Qwen/Phi
order, CommonCorpus has 358,024/364,800/353,912 valid domain tokens and
PMC has 206,908/213,554/202,637. InstructReplay additionally uses 9,600
sequences with 3,485,818/3,375,216/2,996,543 valid replay tokens. The
corresponding replay input-context token counts are
2,469,551/1,955,081/1,826,685.

\section{Paired Uncertainty and Comparative
Summaries}\label{app:stats}

Domain questions are joined when they share a source or fact identifier.
The resulting connected components number 386 for CommonCorpus and 440
for PMC. Bootstrap draws sample these clusters from all 512 recorded
questions, using common indices across all nine methods in a
model--domain setting. Each question carries its eligibility status and
observed confirmation label into every replicate; the 138 ineligible
CommonCorpus questions contribute zero confirmations, preserving the
recorded evaluation population. General resampling is stratified by MMLU
subject, MATH-500 difficulty level, and BBH task, with IFEval and
HumanEval+ each treated as one stratum. Every one of 200,000 replicates recomputes A, F, W, and S.
If cluster \(c\) contains \(n_c\) questions and is drawn \(m_c\) times,
the replicate confirmation rate is \(\sum_c m_c C_c/\sum_c m_c n_c\);
unequal cluster sizes therefore retain question weighting, with a
resampled denominator.

We report two interval constructions. The main text
(Table~\ref{tab:robust}) uses percentile intervals of each pairwise
difference in S, Bonferroni-corrected over all 36 pairs among the eight methods and M0 in a setting. These intervals target the reported quantity directly. Of the 42 InstructReplay--baseline intervals, 37 exclude zero; the other five are Self-Rewrite in the three CommonCorpus settings and RAFT on CommonCorpus Qwen3 and Phi-4-mini. Because positive-part
and maximum operations make F and W nonsmooth at zero-decline
boundaries, we also report a more conservative construction. It forms a 95\% max-T simultaneous band over 36 pairwise Domain differences and 40 signed General changes relative to M0 within a panel and maps that band conservatively to the reported differences. The band covers all pairs,
including the subsequently displayed closest baseline. The
implementation standardizes each centered bootstrap component by its
empirical bootstrap standard deviation (sample SD over all draws).
Components with SD at most \(10^{-12}\) are omitted from the maximum.
The 95th percentile of that maximum absolute standardized deviation
supplies a common cutoff, multiplied by each component SD to form its
band. Each component uses one SD estimated across the bootstrap draws.
For a signed general change bounded by \([l,u]\), its drop lies in
\([\max(0,-u),\max(0,-l)]\); averaging or maximizing these bounds gives
F or W bounds. A pair's S interval subtracts the upper/lower F
difference bounds from the corresponding acquisition bounds.
Zero-variance components have zero band width. Each of the six
model--domain settings forms a separate comparison family. The family
contains all 36 unordered pairs of its nine methods and 40 signed
General changes. Under this construction, 13 of the 42
InstructReplay--baseline intervals exclude zero, none of them against
the closest baseline of a setting.

\textbf{Replay controls.} BOS replay, Hard and Gold replay are outside the 36-pair family. Table~\ref{tab:control-ci} gives paired percentile intervals for InstructReplay minus each control in $S$, with the same resampling design and 200{,}000 draws and a separate Bonferroni correction over the three controls in each setting. The interval excludes zero against Gold replay in two settings (PMC SmolLM3 and Phi-4-mini), against BOS replay in four (CommonCorpus Qwen3 and Phi-4-mini, and PMC Qwen3 and Phi-4-mini) and against Hard in none. If the three pairs are instead added to the 36-pair family (Bonferroni over 39), intervals exclude zero against Gold replay in PMC Phi-4-mini and against BOS replay in CommonCorpus Qwen3 and Phi-4-mini, and PMC Phi-4-mini. These intervals resample questions only; Appendix~\ref{app:scoring} bounds the effect of unresolved judge labels on the CommonCorpus comparisons.

\begin{table*}[!tbp]
\centering\small
\setlength{\tabcolsep}{4pt}
\begin{tabular*}{\textwidth}{@{\extracolsep{\fill}}lrrrrrr@{}}
\toprule
 & \multicolumn{2}{c}{vs.\ Gold replay} & \multicolumn{2}{c}{vs.\ Hard} & \multicolumn{2}{c}{vs.\ BOS replay} \\
\cmidrule(lr){2-3}\cmidrule(lr){4-5}\cmidrule(lr){6-7}
Setting & $\Delta S$ & 95\% CI & $\Delta S$ & 95\% CI & $\Delta S$ & 95\% CI \\
\midrule
CommonCorpus SmolLM3 & 4.51 & [$-0.44$, 9.06] & $-0.88$ & [$-4.11$, 2.27] & 1.28 & [$-2.40$, 5.01] \\
CommonCorpus Qwen3 & 2.47 & [$-2.71$, 7.53] & 0.96 & [$-3.31$, 5.05] & 9.05$^\ast$ & [3.73, 13.70] \\
CommonCorpus Phi-4-mini & 4.10 & [$-1.03$, 9.17] & 2.22 & [$-1.30$, 6.05] & 9.94$^\ast$ & [4.86, 14.86] \\
PMC SmolLM3 & 6.23$^\ast$ & [1.38, 10.58] & 1.35 & [$-2.34$, 4.46] & 2.99 & [$-0.79$, 6.20] \\
PMC Qwen3 & 2.06 & [$-1.73$, 6.31] & 2.29 & [$-1.80$, 5.95] & 5.64$^\ast$ & [0.72, 10.08] \\
PMC Phi-4-mini & 9.60$^\ast$ & [4.23, 14.45] & 0.92 & [$-2.91$, 4.65] & 11.05$^\ast$ & [5.21, 15.75] \\
\bottomrule
\end{tabular*}
\caption{InstructReplay minus each replay control in $S$ (pp). Intervals are paired cluster-bootstrap percentile intervals, Bonferroni-corrected over the three controls in each setting; the main 36-pair family of Table~\ref{tab:robust} is unchanged. $^\ast$The interval excludes zero.}
\label{tab:control-ci}
\end{table*}

All 42 InstructReplay--baseline point differences accompany the source
data. The Pareto count in Section~\ref{sec:main} uses only the frozen A and F
values: a baseline point is dominated when InstructReplay has at least
as much acquisition and at most as much forgetting, with one strict
improvement. The counts by model are 4/5/6 on CommonCorpus and 7/7/6 on
PMC. The intervals quantify question-sampling variation conditional on the
recorded checkpoints and judge votes.

The choice to penalize declines without cancelling them against
improvements affects ranking. Under the alternative utility
\(U=A+\frac{1}{K}\sum_{k=1}^{K}(G_k-G_k^0)\), InstructReplay remains
highest in all six settings; the smallest margin is 0.55 points over
Self-Rewrite on CommonCorpus Qwen.

\textbf{Weighting acquisition against forgetting.} For $S_\beta=A-\beta F$, InstructReplay beats a baseline with higher acquisition whenever $\beta$ exceeds the ratio of their acquisition difference to their forgetting difference. Because InstructReplay has the lowest $F$ in every setting, it beats every baseline with lower or equal acquisition for all $\beta>0$. Table~\ref{tab:beta} gives, for each setting, the smallest $\beta$ above which InstructReplay has the highest $S_\beta$ among the seven Table~\ref{tab:results-1} baselines, and the baseline that determines it.

\begin{table}[!htbp]
\centering\small
\setlength{\tabcolsep}{4pt}
\begin{tabular*}{\columnwidth}{@{\extracolsep{\fill}}llrl@{}}
\toprule
Domain & Model & $\beta^\ast$ & Binding baseline \\
\midrule
CommonCorpus & SmolLM3-3B & 0.22 & RAFT \\
CommonCorpus & Qwen3-4B & 0.31 & RAFT \\
CommonCorpus & Phi-4-mini & 0.25 & RAFT \\
PMC & SmolLM3-3B & 0 & -- \\
PMC & Qwen3-4B & 0 & -- \\
PMC & Phi-4-mini & 0.10 & RAFT \\
\bottomrule
\end{tabular*}
\caption{Smallest weight $\beta^\ast$ above which InstructReplay has the highest $S_\beta=A-\beta F$ among the seven baselines. A value of 0 means it is highest for every $\beta>0$.}
\label{tab:beta}
\end{table}

\section{Variation Across Training Runs}\label{app:runs}

InstructReplay and the two strongest baselines, Self-Rewrite and RAFT, were trained three times in every setting. The runs differ in training randomness and in the order of the training pairs, with each domain sequence kept paired with its replay sequence, and they share the offline supervision: InstructReplay's saved responses, the Self-Rewrite answers and the RAFT targets. Every run finishes 300 optimizer steps and is evaluated with the same questions, judges and general subsets; the main text reports the first run of each method. Table~\ref{tab:runs-s} gives $S$ as mean $\pm$ sample standard deviation over the three runs together with InstructReplay's advantage computed run by run, and Table~\ref{tab:runs-af} decomposes it into $A$ and $F$.

InstructReplay has the highest mean $S$ in every setting and exceeds both baselines in 35 of the 36 run-by-run comparisons; the exception is run 3 on CommonCorpus Qwen3, where Self-Rewrite leads by 0.23 points. The standard deviation of $S$ across runs ranges from 0.64 to 2.42 points for InstructReplay, 0.35 to 2.92 for Self-Rewrite and 0.66 to 5.29 for RAFT.
InstructReplay's MMLU change also varies across runs: $-4.56$, $-4.56$ and $-0.35$ points on CommonCorpus SmolLM3 and $-3.16$, $-0.70$ and $-0.35$ on PMC Phi-4-mini. Its mean over all 18 runs is $-1.05$, against $+0.55$ for Self-Rewrite and $+0.82$ for RAFT; per-run values are in the supplement.

\begin{table*}[!tbp]
\centering\small
\setlength{\tabcolsep}{3pt}
\begin{tabular*}{\textwidth}{@{\extracolsep{\fill}}lrrrrrrr@{}}
\toprule
 & \multicolumn{3}{c}{$S$, mean $\pm$ SD over three runs (pp)} & \multicolumn{2}{c}{IR $-$ Self-Rewrite} & \multicolumn{2}{c}{IR $-$ RAFT} \\
\cmidrule(lr){2-4}\cmidrule(lr){5-6}\cmidrule(lr){7-8}
Model & InstructReplay & Self-Rewrite & RAFT & mean & min & mean & min \\
\midrule
\multicolumn{8}{l}{\textit{CommonCorpus}} \\
\addlinespace[2pt]
SmolLM3-3B & {\boldmath\(2.24\pm1.07\)} & \underline{\(-6.09\pm0.35\)} & \(-11.70\pm1.91\) & 8.33 & 6.84 & 13.93 & 11.24 \\
Qwen3-4B & {\boldmath\(5.84\pm2.36\)} & \underline{\(5.30\pm1.32\)} & \(4.02\pm2.48\) & 0.54 & $-0.23$ & 1.82 & 1.23 \\
Phi-4-mini & {\boldmath\(6.81\pm0.64\)} & \underline{\(0.34\pm1.52\)} & \(0.08\pm0.66\) & 6.47 & 5.16 & 6.73 & 6.16 \\
\midrule
\multicolumn{8}{l}{\textit{PMC}} \\
\addlinespace[2pt]
SmolLM3-3B & {\boldmath\(18.69\pm2.00\)} & \(-8.92\pm2.82\) & \underline{\(-4.42\pm1.97\)} & 27.61 & 25.88 & 23.11 & 19.17 \\
Qwen3-4B & {\boldmath\(17.07\pm2.42\)} & \underline{\(8.11\pm2.03\)} & \(7.56\pm4.54\) & 8.96 & 8.35 & 9.52 & 7.80 \\
Phi-4-mini & {\boldmath\(26.24\pm1.39\)} & \(7.74\pm2.92\) & \underline{\(10.41\pm5.29\)} & 18.50 & 15.60 & 15.83 & 12.74 \\
\bottomrule
\end{tabular*}
\caption{Acquisition--retention score $S$ over three training runs of InstructReplay (IR), Self-Rewrite and RAFT (pp). Differences are computed between runs with the same index, so each setting contributes three per baseline; min is the smallest of the three. In each row, bold marks the highest and underlining the second-highest mean $S$. Ties at the shown precision share marks.}
\label{tab:runs-s}
\end{table*}

\begin{table*}[!tbp]
\centering\small
\setlength{\tabcolsep}{3pt}
\begin{tabular*}{\textwidth}{@{\extracolsep{\fill}}lrrrrrr@{}}
\toprule
 & \multicolumn{3}{c}{$A\uparrow$, mean $\pm$ SD (pp)} & \multicolumn{3}{c}{$F\downarrow$, mean $\pm$ SD (pp)} \\
\cmidrule(lr){2-4}\cmidrule(lr){5-7}
Model & InstructReplay & Self-Rewrite & RAFT & InstructReplay & Self-Rewrite & RAFT \\
\midrule
\multicolumn{7}{l}{\textit{CommonCorpus}} \\
\addlinespace[2pt]
SmolLM3-3B & \underline{\(5.79\pm1.18\)} & \(5.34\pm0.63\) & {\boldmath\(8.53\pm0.30\)} & {\boldmath\(3.56\pm1.61\)} & \underline{\(11.43\pm0.56\)} & \(20.23\pm1.61\) \\
Qwen3-4B & \(9.31\pm0.60\) & \underline{\(9.83\pm0.79\)} & {\boldmath\(10.29\pm0.41\)} & {\boldmath\(3.47\pm1.81\)} & \underline{\(4.53\pm0.79\)} & \(6.27\pm2.12\) \\
Phi-4-mini & \underline{\(9.24\pm0.63\)} & \(8.01\pm0.78\) & {\boldmath\(11.00\pm1.08\)} & {\boldmath\(2.43\pm0.69\)} & \underline{\(7.67\pm0.82\)} & \(10.92\pm0.77\) \\
\midrule
\multicolumn{7}{l}{\textit{PMC}} \\
\addlinespace[2pt]
SmolLM3-3B & {\boldmath\(22.85\pm1.98\)} & \(15.62\pm0.52\) & \underline{\(21.55\pm1.48\)} & {\boldmath\(4.16\pm0.09\)} & \underline{\(24.55\pm2.96\)} & \(25.96\pm0.50\) \\
Qwen3-4B & \underline{\(17.58\pm2.03\)} & \underline{\(17.58\pm1.41\)} & {\boldmath\(18.03\pm1.47\)} & {\boldmath\(0.51\pm0.40\)} & \underline{\(9.47\pm1.26\)} & \(10.48\pm3.89\) \\
Phi-4-mini & {\boldmath\(27.73\pm1.41\)} & \(17.90\pm1.18\) & \underline{\(27.21\pm3.43\)} & {\boldmath\(1.50\pm0.22\)} & \underline{\(10.17\pm1.75\)} & \(16.81\pm1.98\) \\
\bottomrule
\end{tabular*}
\caption{Acquisition $A$ and forgetting $F$ over the same three training runs (pp). In each row and metric, bold marks the best and underlining the second-best mean (highest $A$, lowest $F$). Ties at the shown precision share marks.}
\label{tab:runs-af}
\end{table*}

\section{Replay Controls: BOS Context, Hard Targets and
Gold Answers}\label{app:replay}

\begin{figure*}[!tbp]
\centering
\begin{tikzpicture}
\begin{groupplot}[group style={group size=3 by 1, horizontal sep=0.55cm, yticklabels at=edge left},
  width=0.33\textwidth, height=6.2cm, ymin=-0.6, ymax=5.6, y dir=reverse, ytick={0,...,5},
  yticklabels={CC/Smol,CC/Qwen,CC/Phi,PMC/Smol,PMC/Qwen,PMC/Phi}, ticklabel style={font=\scriptsize},
  title style={font=\small}, axis lines*=left, xmajorgrids, grid style={gray!20}]
\nextgroupplot[title={$\Delta A$ (pp) $\uparrow$}, xmin=-3, xmax=3,
  legend to name=rdlegend, legend columns=2, legend style={font=\scriptsize, draw=none, /tikz/every even column/.append style={column sep=8pt}}]
\draw[densely dashed, gray] (axis cs:0,-0.6) -- (axis cs:0,5.6);
\addplot[only marks, mark=diamond, hardcol, very thick, mark size=2.6pt] table[x=Ahard, y=y] {figures/data/replaydiff.dat};
\addplot[only marks, mark=square, boscol, thick, mark size=2.1pt] table[x=Abos, y=y] {figures/data/replaydiff.dat};
\legend{Hard, BOS replay}
\nextgroupplot[title={$\Delta F$ (pp) $\downarrow$}, xmin=-14, xmax=3]
\draw[densely dashed, gray] (axis cs:0,-0.6) -- (axis cs:0,5.6);
\addplot[only marks, mark=diamond, hardcol, very thick, mark size=2.6pt] table[x=Fhard, y=y] {figures/data/replaydiff.dat};
\addplot[only marks, mark=square, boscol, thick, mark size=2.1pt] table[x=Fbos, y=y] {figures/data/replaydiff.dat};
\nextgroupplot[title={$\Delta S$ (pp) $\uparrow$}, xmin=-6, xmax=17,
  nodes near coords, point meta=explicit symbolic, every node near coord/.append style={font=\tiny, anchor=west, xshift=3pt}]
\draw[densely dashed, gray] (axis cs:0,-0.6) -- (axis cs:0,5.6);
\addplot[only marks, mark=diamond, hardcol, very thick, mark size=2.6pt] table[x=Shard, y=yh, meta=Shardlab] {figures/data/replaydiff.dat};
\addplot[only marks, mark=square, boscol, thick, mark size=2.1pt] table[x=Sbos, y=yb, meta=Sboslab] {figures/data/replaydiff.dat};
\end{groupplot}
\end{tikzpicture}

\ref{rdlegend}
\caption{InstructReplay minus Hard and BOS replay across all six settings (pp). Acquisition, forgetting, and score differences occupy separate columns; arrows indicate the favorable direction and zero lines mark ties. The score column labels every value. CC abbreviates CommonCorpus.}
\label{fig:replaydiff}
\end{figure*}
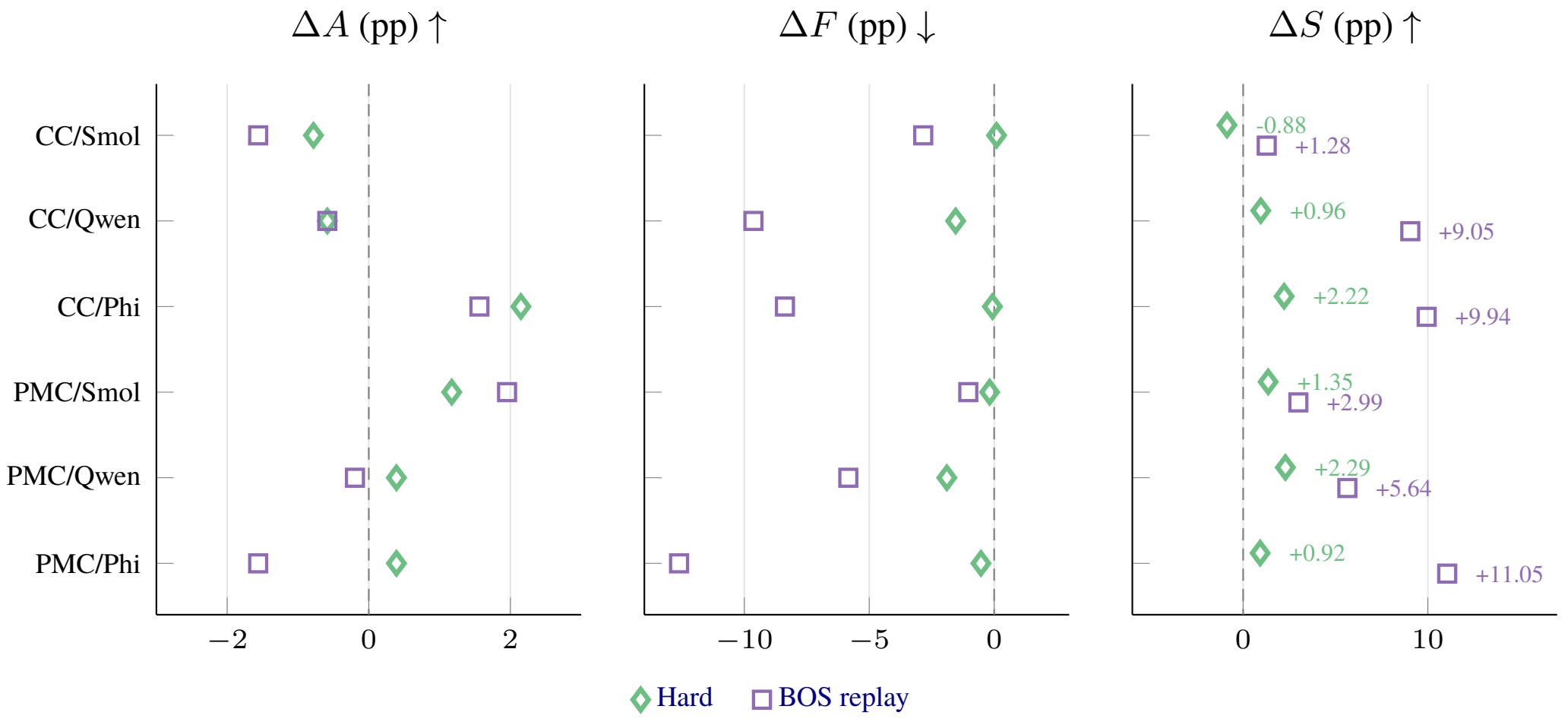

InstructReplay and Hard share the saved instruction-response prefixes,
with frozen-distribution KL and saved-token cross-entropy targets,
respectively. BOS replay uses BOS-conditioned prefixes with the same KL.
Domain-KL instead uses domain gold-answer prefixes (Appendix~\ref{app:training}).
Gold replay keeps InstructReplay's instruction prompts token for token but
replaces $M_0$'s responses with the collection's own final assistant
answers, truncated to 512 tokens and closed with the model's end-of-turn token. This gives about 2.8M response tokens per model, against 3.0M to 3.5M for InstructReplay. It trains on them with cross-entropy and otherwise uses
the trainer, coefficient, ratio and schedule of Hard, so Hard and Gold
replay differ only in who wrote the replayed answers.
Figure~\ref{fig:replaydiff} decomposes the measured differences for BOS
replay and Hard; Table~\ref{tab:results-18} reports all four variants.

On SmolLM3, Gold replay lowers MMLU by 18.2 and 12.3 points, far more than
any other method. A probe shows that SmolLM3 takes on formatting
from the dataset's answers; the scorer is identical for all methods. Of the 1,920 academic answers, 986 open with a literal empty reasoning block, \texttt{<think></think>}, which for SmolLM3 and
Qwen3 repeats the block that the chat template has already placed in the
prompt; no response sampled from $M_0$ does. After Gold replay, SmolLM3 assigns probability 0.29 to 0.32 to opening its turn with this token on the 285 MMLU prompts, against 0.00 for $M_0$ and Hard. At the scored position the four options keep 0.83 and 0.84 of the probability mass, against 0.99 for $M_0$ and 0.98 for Hard. The remainder goes mainly to Markdown emphasis and end-of-turn tokens, which the dataset's answers use and $M_0$'s rarely do.
Because MMLU takes the most likely of the four options at that position,
the leaked mass does not by itself change a prediction. To test the
artifact directly, we removed the leading block from the 986 answers,
rebuilt the SmolLM3 buffer otherwise unchanged and retrained both SmolLM3
cells, evaluating General ability only. The opening probability falls to 0.00, but MMLU does not recover:
37.89 on CommonCorpus and 47.72 on PMC, against 41.05 and 47.02 with the
block and 59.30 for $M_0$. The options keep 0.87 and 0.84 of the
probability mass, the remainder still going mainly to Markdown emphasis.
IFEval is essentially unchanged at 57.50 and 65.83, whereas MATH-500 falls
to 53.00 in both settings and BBH to 61.11 and 53.70, so $F$ rises to 15.25
and 14.98. Removing the empty block leaves MMLU below $M_0$ in both
SmolLM3 settings. The remaining loss is unresolved: this intervention
does not separate other formatting effects, answer length, target quality
and knowledge changes.

BOS replay and the hard-target ablation start from the same exact M0 as their InstructReplay counterpart. Domain token IDs, pair order,
effective batch, constant learning rate, and 300-step endpoint are
preserved. Both losses use response-only masks and per-sequence means.
The training records confirm the schedule and zero logit
discrepancy after saving and reloading the final model. Pre-run checks
verify finite nonzero domain gradients and the reference identity.

The hard-target ablation uses \(\mathcal L_D+0.25\mathcal L_{\mathrm{hard}}\), where
\(\mathcal L_{\mathrm{hard}}\) is next-token cross-entropy on the saved instruction responses. A frozen-teacher forward is
retained for diagnostics but excluded from the optimized hard loss. Its
measured compute therefore includes work that a production hard-replay
implementation could omit.

BOS replay uses the BOS-conditioned
sampling of \citet{marek2026forgetting} with the forward KL,
coefficient and schedule of InstructReplay. The model-config BOS token initializes each sequence.
Sampling uses temperature 1, top-p 1, and top-k 0, with EOS stopping
disabled. Each row uses a fixed random state; the generated sequence
is trimmed to that row's instruction-response budget. The
stored 9,600-row buffer is reused across the two domains for each model.
This matches sequence counts and retained response positions per
sequence; content, instruction context, padding and runtime depend on
the replay input. The author implementation averages over valid tokens
and generates replay in successive frozen-reference buffers. Our
adaptation uses sequence means and a precomputed buffer, with per-row
length matching and the common 1:1, 0.25, 300-step schedule. The
authors' instruction-model configuration instead uses a 4:1
domain/replay sequence ratio, five epochs, and a coefficient sweep over
1, 3, and 10.

\begin{table*}[!tbp]
\centering
\footnotesize
\setlength{\tabcolsep}{3pt}
\begin{tabularx}{\textwidth}{@{}l*{10}{>{\raggedleft\arraybackslash}X}@{}}
\toprule
 & \multicolumn{1}{c}{Domain (\%)} & \multicolumn{5}{c}{General (\%)} & \multicolumn{4}{c}{Acquisition / forgetting (pp)} \\
\cmidrule(lr){2-2}\cmidrule(lr){3-7}\cmidrule(lr){8-11}
Replay configuration & Core $\uparrow$ & IF & MM & MA & BBH & HE & $A\uparrow$ & $F\downarrow$ & $W\downarrow$ & $S\uparrow$ \\
\midrule
\multicolumn{11}{l}{\textit{CommonCorpus --- SmolLM3-3B}} \\
\addlinespace[2pt]
\fullrowcolor{oursrow}{\textwidth}InstructReplay & 15.43 & \textbf{67.50} & 54.74 & \textbf{74.00} & \textbf{69.26} & \underline{40.62} & 5.66 & \underline{4.66} & \underline{10.94} & \underline{1.01} \\
Hard & 16.21 & \underline{64.17} & \underline{55.79} & \underline{72.00} & \underline{68.89} & \textbf{43.75} & 6.45 & \textbf{4.56} & \textbf{7.81} & \textbf{1.88} \\
BOS replay & \underline{16.99} & 55.00 & \textbf{57.19} & \underline{72.00} & 66.67 & 39.06 & \underline{7.23} & 7.50 & 15.83 & -0.27 \\
Gold replay & \textbf{17.77} & 58.33 & 41.05 & 62.00 & 63.70 & \textbf{43.75} & \textbf{8.01} & 11.51 & 18.25 & -3.50 \\
\midrule
\multicolumn{11}{l}{\textit{CommonCorpus --- Qwen3-4B}} \\
\addlinespace[2pt]
\fullrowcolor{oursrow}{\textwidth}InstructReplay & 17.38 & \underline{74.17} & 71.23 & \textbf{70.00} & 70.37 & \textbf{62.50} & 9.96 & \textbf{1.40} & \textbf{6.67} & \textbf{8.56} \\
Hard & \underline{17.97} & \textbf{75.83} & 71.23 & \underline{69.00} & \underline{73.33} & \underline{50.00} & \underline{10.55} & \underline{2.95} & \underline{9.38} & \underline{7.60} \\
BOS replay & \underline{17.97} & 55.00 & \textbf{74.04} & \textbf{70.00} & 60.37 & 35.94 & \underline{10.55} & 11.04 & 25.83 & -0.49 \\
Gold replay & \textbf{18.55} & 71.67 & \underline{72.63} & 65.00 & \textbf{75.19} & 45.31 & \textbf{11.13} & 5.05 & 14.06 & 6.09 \\
\midrule
\multicolumn{11}{l}{\textit{CommonCorpus --- Phi-4-mini}} \\
\addlinespace[2pt]
\fullrowcolor{oursrow}{\textwidth}InstructReplay & \textbf{16.41} & \textbf{67.50} & \underline{70.18} & \underline{59.00} & \textbf{72.22} & \textbf{62.50} & \textbf{9.96} & \textbf{2.74} & \underline{10.00} & \textbf{7.22} \\
Hard & 14.26 & \underline{64.17} & \textbf{70.53} & \textbf{62.00} & \underline{71.85} & \underline{57.81} & 7.81 & \underline{2.81} & \textbf{7.00} & \underline{5.00} \\
BOS replay & 14.84 & 50.83 & 69.47 & 44.00 & 67.41 & 51.56 & 8.40 & 11.12 & 25.00 & -2.72 \\
Gold replay & \underline{16.21} & 60.83 & 69.82 & \textbf{62.00} & 62.96 & 50.00 & \underline{9.77} & 6.65 & 10.74 & 3.12 \\
\midrule
\multicolumn{11}{l}{\textit{PMC --- SmolLM3-3B}} \\
\addlinespace[2pt]
\fullrowcolor{oursrow}{\textwidth}InstructReplay & \textbf{38.67} & \textbf{67.50} & \underline{59.65} & \underline{71.00} & \underline{66.67} & \textbf{40.62} & \textbf{22.46} & \textbf{4.26} & \textbf{10.94} & \textbf{18.20} \\
Hard & 37.50 & \underline{66.67} & \textbf{60.70} & \textbf{74.00} & \textbf{68.15} & \underline{39.06} & 21.29 & \underline{4.44} & \underline{12.50} & \underline{16.84} \\
BOS replay & 36.72 & 60.83 & \underline{59.65} & \textbf{74.00} & \textbf{68.15} & \textbf{40.62} & 20.51 & 5.30 & \textbf{10.94} & 15.21 \\
Gold replay & \underline{38.48} & 65.83 & 47.02 & 68.00 & 58.15 & 35.94 & \underline{22.27} & 10.29 & 15.62 & 11.97 \\
\midrule
\multicolumn{11}{l}{\textit{PMC --- Qwen3-4B}} \\
\addlinespace[2pt]
\fullrowcolor{oursrow}{\textwidth}InstructReplay & \underline{37.30} & \textbf{80.00} & \textbf{72.63} & \underline{70.00} & \underline{74.07} & \textbf{62.50} & \underline{18.75} & \textbf{0.17} & \textbf{0.83} & \textbf{18.58} \\
Hard & 36.91 & \underline{78.33} & 71.93 & \underline{70.00} & \textbf{77.78} & 51.56 & 18.36 & 2.06 & \underline{7.81} & 16.30 \\
BOS replay & \textbf{37.50} & 74.17 & \underline{72.28} & \textbf{72.00} & 64.81 & 37.50 & \textbf{18.95} & 6.00 & 21.88 & 12.94 \\
Gold replay & 36.91 & 71.67 & \textbf{72.63} & 69.00 & 70.00 & \underline{59.38} & 18.36 & \underline{1.83} & 9.17 & \underline{16.53} \\
\midrule
\multicolumn{11}{l}{\textit{PMC --- Phi-4-mini}} \\
\addlinespace[2pt]
\fullrowcolor{oursrow}{\textwidth}InstructReplay & \underline{37.30} & \textbf{67.50} & 68.42 & \textbf{69.00} & \underline{70.74} & \textbf{59.38} & \underline{26.17} & \textbf{1.39} & \textbf{3.16} & \textbf{24.78} \\
Hard & 36.91 & \underline{66.67} & \underline{69.12} & \underline{65.00} & \textbf{72.22} & \underline{57.81} & 25.78 & \underline{1.92} & \underline{4.00} & \underline{23.86} \\
BOS replay & \textbf{38.87} & 59.17 & 67.72 & 45.00 & 62.59 & 34.38 & \textbf{27.73} & 14.00 & 24.00 & 13.73 \\
Gold replay & 35.94 & \textbf{67.50} & \textbf{69.82} & 62.00 & 57.04 & 34.38 & 24.80 & 9.63 & 21.88 & 15.18 \\
\bottomrule
\end{tabularx}
\caption{Direct comparison of InstructReplay, Hard, BOS replay and Gold replay in all six settings. InstructReplay rows are reproduced from Tables \ref{tab:cc-full}--\ref{tab:pmc-full}. Within each four-row group, bold/underline mark the best/second-best value in every metric, with directions and ties as in Table~\ref{tab:cc-full}; Hard has the highest $S$ in one setting. IF, MM, MA and HE denote IFEval, MMLU, MATH-500 and HumanEval+. Domain reports \(100C/N_D\) (\%) with \(N_D=512\); unknowns are retained in the denominator. General percentages use the same scorers as Tables~\ref{tab:cc-full}--\ref{tab:pmc-full}; A/F/W/S are in pp. Shaded rows mark InstructReplay.}
\label{tab:results-18}
\end{table*}

\textbf{Hard-target ablation results.} InstructReplay leads Hard in five
settings; Hard leads by 0.88 pp on CommonCorpus Smol. On PMC,
InstructReplay leads by 0.92 to 2.29 pp.~CommonCorpus Phi's 2.22-pp
advantage mostly reflects acquisition (2.15 pp), with only 0.07 pp less
average forgetting. The comparison
uses the fixed coefficient 0.25 for both losses, whose gradient
magnitudes can differ. The balance between acquisition and forgetting
favors different target forms across these settings.

\begin{table*}[!tbp]
\centering
\small
\setlength{\tabcolsep}{3pt}
\begin{tabular*}{\textwidth}{@{\extracolsep{\fill}}lrrrrr@{}}
\toprule
 & & \multicolumn{2}{c}{CommonCorpus (GPUh)} & \multicolumn{2}{c}{PMC (GPUh)} \\
\cmidrule(lr){3-4}\cmidrule(lr){5-6}
Model & Replay positions & Hard & BOS replay & Hard & BOS replay \\
\midrule
SmolLM3-3B & 3,485,818 & 1.57 & \textbf{1.50} & 1.58 & \textbf{1.50} \\
Qwen3-4B & 3,375,216 & 2.24 & \textbf{1.89} & 2.20 & \textbf{1.90} \\
Phi-4-mini & 2,996,543 & 1.98 & \textbf{1.52} & 1.72 & \textbf{1.45} \\
\bottomrule
\end{tabular*}
\caption{Shared retained-token budgets and per-domain training costs (GPUh) for the additional replay baseline and hard-target ablation. All 12 runs use eight A100 40-GB devices. Each model uses the same replay budget in both domains, matching its InstructReplay response budget; GPU-hours include the diagnostic teacher forward in Hard. Within each model and domain, bold marks the lower recorded GPU-hours.}
\label{tab:results-19}
\end{table*}

The 12 training runs consume 21.03 allocated GPU-hours in total. Separate BOS buffer generation consumes 5.26 allocated
GPU-hours including startup: 1.65 for Smol on A800, 2.03 for Qwen on
A800, and 1.58 for Phi on A100. Generation produces 4,915,200 tokens per
model before row-wise trimming; retained counts are 3,485,818,
3,375,216, and 2,996,543. These one-time buffer costs are not charged
twice for reuse across domains. Smoke runs, evaluation and failed
attempts are outside the listed training costs.

\section{Reproducibility
Materials}\label{app:repro}

The arXiv ancillary files provide result tables for all 114 evaluated endpoints, hash-anonymized domain labels and the scoring materials. The standalone supplement also includes the LaTeX source. The result tables give the five General scores, the domain confirmation count and $A$, $F$ and $S$ of every endpoint, together with the paired intervals, output lengths and training costs. The labels cover 102 endpoints: the 54 rows of Table~\ref{tab:results-1}, the 36 additional training runs and the 12 BOS replay and Hard runs. For each endpoint they give the three judge votes and the majority label on all 512 questions, and the CommonCorpus eligibility mask is included. The scoring materials contain the data-construction and judge prompts, the data cards, the source examples of Appendix~\ref{app:data}, saved short General outputs and the BBH answer extractor.

A verification script recomputes every labeled endpoint's confirmation count from the judge votes, the $A$, $F$ and $S$ values from the scores, the confirmation transitions and the unknown-only ranges of Appendix~\ref{app:scoring}. It uses the Python standard library only. Released examples include two attributed CC BY 4.0 source excerpts with example question--answer pairs, short General outputs and six complete BBH response examples. Full source and evaluation corpora are withheld because row-level redistribution permission is not established for the complete data mixture.

All models are public. Student
repositories are:

\begin{itemize}
\tightlist
\item
  \nolinkurl{HuggingFaceTB/SmolLM3-3B}
\item
  \nolinkurl{Qwen/Qwen3-4B}
\item
  \nolinkurl{microsoft/Phi-4-mini-instruct}
\end{itemize}

Judge and fusion repositories are:

\begin{itemize}
\tightlist
\item
  \nolinkurl{Qwen/Qwen3.8-27B}
\item
  \nolinkurl{google/gemma-4-31B-it}
\item
  \nolinkurl{openai/gpt-oss-20b}
\end{itemize}

The recorded training runtime uses Transformers 4.57.1; judge deployment
logs report vLLM 0.19.0.

\section{Output-Length Analysis}\label{app:length}

\begin{figure*}[!tbp]
\centering
\begin{tikzpicture}
\begin{groupplot}[group style={group size=3 by 1, horizontal sep=0.9cm},
  width=0.34\textwidth, height=4.6cm, xmin=0, xmax=1.35, ymin=-75, ymax=12,
  xlabel={Median length / $M_0$}, xlabel style={font=\small},
  ticklabel style={font=\scriptsize}, title style={font=\small\bfseries},
  length classes,
  axis lines*=left, ymajorgrids, grid style={gray!20}]
\nextgroupplot[title=IFEval, ylabel={$\Delta$ accuracy vs.\ $M_0$ (pp)}, ylabel style={font=\small},
  legend to name=lenlegend, legend columns=6, legend cell align=left, legend style={font=\scriptsize, draw=none, /tikz/every even column/.append style={column sep=6pt}}]
\addplot[scatter, only marks, scatter src=explicit symbolic] table[x=ratio, y=delta, meta=cls] {figures/data/length-IFEval.dat};
\legend{Other baselines, Domain-KL, BOS replay, Hard, InstructReplay, Naive SFT}
\nextgroupplot[title=MATH-500]
\addplot[scatter, only marks, scatter src=explicit symbolic] table[x=ratio, y=delta, meta=cls] {figures/data/length-MATH.dat};
\nextgroupplot[title=BBH]
\addplot[scatter, only marks, scatter src=explicit symbolic] table[x=ratio, y=delta, meta=cls] {figures/data/length-BBH.dat};
\end{groupplot}
\end{tikzpicture}

\ref{lenlegend}
\caption{Accuracy change versus median response length relative to $M_0$ for all 60 adapted checkpoints, extending Figure~\ref{fig:length-main}a to MATH-500 and BBH. Spearman $\rho$ is 0.77 (IFEval), 0.81 (MATH-500) and 0.84 (BBH).}
\label{fig:length}
\end{figure*}
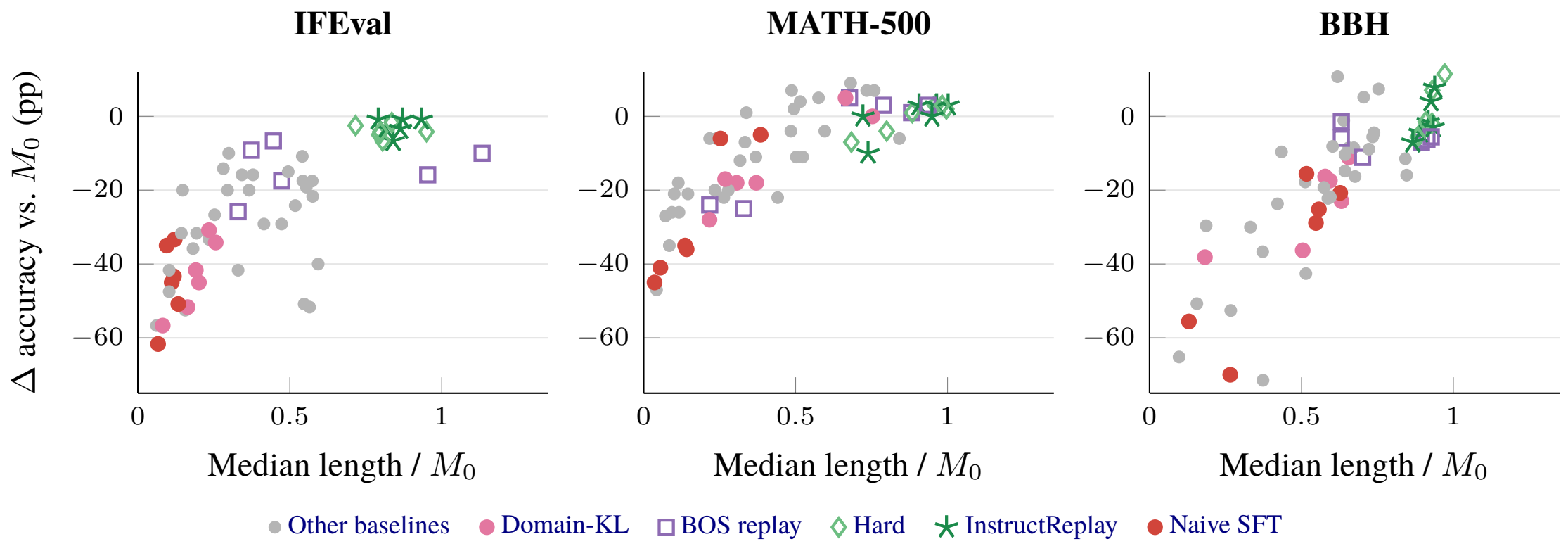

This analysis reads the saved greedy General outputs of all 66 checkpoints (six $M_0$ evaluations, 48 trained methods and 12 replay controls) and counts whitespace-separated words per response. For each checkpoint and generative benchmark we take the median response length and divide it by that of the same model's $M_0$. Table~\ref{tab:length} averages these ratios over the six settings, and Figure~\ref{fig:length} relates them to the accuracy change of every adapted checkpoint. The Spearman correlation over the 60 adapted checkpoints is 0.77 for IFEval, 0.81 for MATH-500 and 0.84 for BBH. The 60 exclude the six Gold replay checkpoints; Table~\ref{tab:length} adds their length ratios, and including them gives correlations of 0.80, 0.73 and 0.74. Gold replay keeps $M_0$'s MATH-500 length (ratio 1.00) yet loses 4.3 points there, which agrees with the question-level result below: departure from $M_0$'s response, not shortening alone, goes with failure. Shorter outputs are not wrong in themselves; the correlation shows that, in this study, the methods that preserve accuracy are those that preserve the initial model's response style.

\begin{table}[!htbp]
\centering\small
\setlength{\tabcolsep}{4pt}
\begin{tabular*}{\columnwidth}{@{\extracolsep{\fill}}lrrr@{}}
\toprule
Method & IFEval & MATH-500 & BBH \\
\midrule
Naive SFT & 0.11 & 0.17 & 0.44 \\
DFT & 0.13 & 0.19 & 0.34 \\
Domain-KL & 0.19 & 0.43 & 0.52 \\
LoRA-SFT & 0.21 & 0.42 & 0.71 \\
SDFT & 0.34 & 0.28 & 0.52 \\
RAFT & 0.52 & 0.36 & 0.59 \\
Self-Rewrite & 0.53 & 0.58 & 0.65 \\
BOS replay & 0.62 & 0.64 & 0.78 \\
Gold replay & 0.74 & \textbf{1.00} & 0.88 \\
Hard & \underline{0.82} & \underline{0.88} & \textbf{0.92} \\
\fullrowcolor{oursrow}{\columnwidth}InstructReplay & \textbf{0.85} & \underline{0.88} & \underline{0.91} \\
\bottomrule
\end{tabular*}
\caption{Median response length relative to $M_0$ (1.00), averaged over six settings. $M_0$'s mean medians are 190 (IFEval), 242 (MATH-500) and 122 (BBH) words. The shaded row is InstructReplay. Bold/underline mark the closest/second-closest ratio to 1.00 per column, including ties.}
\label{tab:length}
\end{table}

\textbf{Instruction categories.} Shorter answers could fail IFEval mechanically, by violating word-count constraints. Table~\ref{tab:ifeval-categories} lists all nine instruction categories. Naive SFT's strict pass rate falls by 11 points on length constraints but by 55 to 77 points on constraints that require structure or specific content, namely format, start/end phrases, content and combinations. A terse answer violates these even when it meets any length requirement. Losses are smaller on letter case, keywords and punctuation, at 20, 31 and 9.5 points; the 67-point loss on response language rests on five instructions per model. InstructReplay stays within a few points of $M_0$ in most categories.

\textbf{Question-level failures.} Table~\ref{tab:item-length} restricts attention to questions that $M_0$ answers correctly and groups each adapted response by its length relative to $M_0$'s response to the same question. Pooled over checkpoints, failure is least frequent when the length stays within 25\% of $M_0$'s and rises both for much shorter and for much longer responses, in all three benchmarks. Within Naive SFT alone, failure falls steeply as responses approach $M_0$'s length. The pooled rows mix methods, so they describe where failures occur rather than isolate a causal effect of length.

\begin{table*}[!tbp]
\centering\small
\setlength{\tabcolsep}{4pt}
\begin{tabular*}{\textwidth}{@{\extracolsep{\fill}}lrrrrrrr@{}}
\toprule
Instruction category & $n$ & $M_0$ & Naive & Dom.-KL & BOS & Hard & IR \\
\midrule
Length (words, sentences, paragraphs) & 32 & 59.4 & 48.4 & 46.4 & \textbf{63.5} & 60.4 & \underline{61.5} \\
Format (sections, lists, JSON) & 36 & 89.8 & 35.2 & 44.9 & 83.3 & \textbf{91.7} & \underline{90.7} \\
Start or end phrase & 13 & 89.7 & 12.8 & 17.9 & 55.1 & \underline{87.2} & \textbf{88.5} \\
Content (placeholders, postscript) & 12 & 100.0 & 23.6 & 41.7 & 83.3 & \textbf{97.2} & \underline{94.4} \\
Combination (repeat prompt, two answers) & 16 & 72.9 & 10.4 & 13.5 & 50.0 & \textbf{62.5} & \underline{60.4} \\
Response language & 5 & 100.0 & 33.3 & 16.7 & 56.7 & \underline{80.0} & \textbf{83.3} \\
Letter case & 23 & 84.1 & 63.8 & 54.3 & 76.8 & \underline{79.7} & \textbf{83.3} \\
Keywords & 44 & 81.8 & 50.8 & 52.3 & 69.7 & \underline{76.1} & \textbf{78.4} \\
Punctuation & 14 & 90.5 & 81.0 & 69.0 & 72.6 & \underline{84.5} & \textbf{86.9} \\
\bottomrule
\end{tabular*}
\caption{IFEval strict pass rate (\%) by instruction category over the six settings. $n$ is the number of instructions per model; $M_0$ is counted once per model and each method once per setting. Dom.-KL: Domain-KL; IR: InstructReplay. Within each category, bold/underline mark the highest/second-highest adapted pass rate; $M_0$ is a reference.}
\label{tab:ifeval-categories}
\end{table*}

\begin{table*}[!tbp]
\centering\small
\setlength{\tabcolsep}{4pt}
\begin{tabular*}{\textwidth}{@{\extracolsep{\fill}}llrrrrr@{}}
\toprule
 & & \multicolumn{5}{c}{Response length relative to $M_0$'s on the same question} \\
\cmidrule(lr){3-7}
Benchmark & Checkpoints & $<0.25$ & 0.25--0.5 & 0.5--0.8 & 0.8--1.25 & $>1.25$ \\
\midrule
IFEval & all adapted & 70 (1464) & 46 (954) & \underline{29} (1108) & \textbf{20} (1295) & 49 (459) \\
 & Naive SFT & 79 (333) & 61 (96) & \textbf{33} (46) & \underline{43} (28) & 52 (25) \\
 & InstructReplay & 17 (6) & \textbf{8} (24) & 14 (129) & \underline{10} (303) & 24 (66) \\
\addlinespace[2pt]
MATH-500 & all adapted & 84 (1131) & 60 (832) & \underline{34} (892) & \textbf{17} (1131) & 37 (154) \\
 & Naive SFT & 93 (274) & 89 (83) & \underline{62} (37) & \textbf{41} (17) & 67 (3) \\
 & InstructReplay & 14 (7) & 26 (27) & \underline{10} (78) & \textbf{7} (276) & 23 (26) \\
\addlinespace[2pt]
BBH & all adapted & 83 (1617) & 58 (1463) & \underline{35} (1809) & \textbf{15} (6265) & 46 (386) \\
 & Naive SFT & 95 (314) & 76 (192) & \underline{55} (165) & \textbf{27} (464) & 74 (19) \\
 & InstructReplay & 20 (10) & \underline{13} (47) & 20 (158) & \textbf{11} (877) & 32 (62) \\
\bottomrule
\end{tabular*}
\caption{Failure rate (\%) of adapted checkpoints on questions that $M_0$ answers correctly, by per-question response length (whitespace words) relative to $M_0$; the number of question--checkpoint pairs is in parentheses. ``All adapted'' pools the 60 adapted checkpoints of Table~\ref{tab:length}. Within each row, bold/underline mark the lowest/second-lowest failure rate across length bins; counts are unranked.}
\label{tab:item-length}
\end{table*}

\end{document}